\documentclass{article}

\usepackage{microtype}
\usepackage{graphicx}
\usepackage[dvipsnames]{xcolor}
\usepackage[tight]{subfigure}
\usepackage{booktabs} % for professional tables

\usepackage{hyperref}

\usepackage[accepted]{icml2024}

\usepackage{amsmath}
\usepackage{amssymb}
\usepackage{mathtools}
\usepackage{amsthm}
\usepackage{bm}
\usepackage{dsfont}
\usepackage{enumitem}

\usepackage[capitalize,noabbrev]{cleveref}

\theoremstyle{plain}
\newtheorem{theorem}{Theorem}[section]

\theoremstyle{definition}
\newtheorem{definition}[theorem]{Definition}
\newtheorem{assumption}[theorem]{Assumption}
\theoremstyle{remark}

\DeclareMathOperator*{\EV}{\mathbb{E}}
\newcommand{\op}{\mathbb{T}}
\newcommand{\U}{\mathcal{U}}
\newcommand{\M}{\mathcal{M}}
\renewcommand{\S}{\mathcal{S}}
\newcommand{\A}{\mathcal{A}}
\renewcommand{\O}{\mathcal{O}}
\renewcommand{\P}{\mathbb{P}}
\newcommand{\Obs}{\mathbb{O}}
\newcommand{\bel}{\bm{b}}
\newcommand{\B}{\mathcal{B}}
\newcommand{\I}{\mathcal{I}}
\newcommand{\Bf}{\mathbb{B}}
\newcommand{\wP}{\widetilde{\mathbb{P}}}
\newcommand{\Tau}{\mathcal{T}}
\newcommand{\1}[1]{\mathds{1}\{#1\}}

\DeclareMathOperator*{\argmax}{arg\,max}

\usepackage[textsize=tiny]{todonotes}

\newcommand\margincomment[4]

\icmltitlerunning{State Entropy Maximization in POMDPs}

\begin{document}

\twocolumn[
\icmltitle{How to Explore with Belief: State Entropy Maximization in POMDPs}
% NEW TITLE
% How to Explore with Belief: State Entropy Maximization in POMDPs



\icmlsetsymbol{equal}{*}

\begin{icmlauthorlist}
\icmlauthor{Riccardo Zamboni}{polimi}
\icmlauthor{Duilio Cirino}{polimi}
\icmlauthor{Marcello Restelli}{polimi}
\icmlauthor{Mirco Mutti}{technion}
\end{icmlauthorlist}

\icmlaffiliation{polimi}{Politecnico di Milano, Milan, Italy.}
\icmlaffiliation{technion}{Technion -- Israel Institute of Technology, Haifa, Israel}

\icmlcorrespondingauthor{Riccardo Zamboni}{riccardo.zamboni@polimi.it}

\icmlkeywords{Reinforcement Learning, Maximum State Entropy Exploration, POMDPs}

\vskip 0.3in
]



\printAffiliationsAndNotice{} % otherwise use the standard text.

\begin{abstract}
Recent works have studied \emph{state entropy maximization} in reinforcement learning, in which the agent's objective is to learn a policy inducing high entropy over states visitation~\cite{hazan2019provably}. They typically assume full observability of the state of the system so that the entropy of the observations is maximized. In practice, the agent may only get \emph{partial} observations, e.g., a robot perceiving the state of a physical space through proximity sensors and cameras. A significant mismatch between the entropy over observations and true states of the system can arise in those settings.
In this paper, we address the problem of entropy maximization over the \emph{true states} with a decision policy conditioned on partial observations \emph{only}. The latter is a generalization of POMDPs, which is intractable in general. We develop a memory and computationally efficient \emph{policy gradient} method to address a first-order relaxation of the objective defined on \emph{belief} states, providing various formal characterizations of approximation gaps, the optimization landscape, and the \emph{hallucination} problem.
This paper aims to generalize state entropy maximization to more realistic domains that meet the challenges of applications.
\end{abstract}

\section{Introduction}
\label{sec:introduction}

The \emph{state entropy maximization} framework, initially proposed in~\citet{hazan2019provably}, is a popular generalization of the Reinforcement Learning~\citep[RL,][]{bertsekas2019reinforcement} problem in which an agent aims to maximize, instead of the cumulative reward, a (self-supervised) objective related to the entropy of the state visitation induced by its policy.

The entropy objective finds motivation as a standalone tool for learning to cover the states of the environment~\cite{hazan2019provably}, a data collection strategy for offline RL~\cite{yarats2022don}, experimental design~\cite{tarbouriech2019active}, or transition model estimation~\cite{tarbouriech2020active, jin2020reward}, and as a surrogate loss for policy pre-training in reward-free settings~\cite{mutti2020ideal}.

While the objective itself is not convex in the policy parameters, it is known to admit a tractable dual formulation~\cite{hazan2019provably}, and several practical methods, also in combination with neural policies, have been developed~\cite{mutti2021task, liu2021behavior, seo2021state, yarats2021reinforcement} as a testament of the promises of the framework for tangible impact on real-world applications.

All of the previous works on state entropy maximization assume the state of the environment is fully observable, such that the agent-environment interaction can be modeled as a Markov Decision Process~\citep[MDP,][]{puterman2014markov}.
Under this assumption, maximizing the entropy over the observations collected from the environment is well-founded. However, the agent may only receive partial observation from the environment in practice. 

Let us think of an autonomous robot for \emph{rescue operations} as an illustrative application: The robot is placed in an unknown terrain with the goal of covering every inch of the ground in order to locate and rescue a wounded human unable to move. The robot cannot access its true position, as well as the human's location-; it can only perceive its surroundings with sensors and cameras. Arguably, maximizing the entropy of the observations, such as changing the camera angle in every direction, is undesirable for the given task. Instead, we would like the robot to maximize the entropy of its position, for which the best policy may entail moving the camera to probe the surroundings and avoid getting stuck, but also to step forward to cover the most ground so that the wounded human can be swiftly located and rescued. 

In this paper, we aim to generalize the state entropy maximization framework to scenarios of the kind of the latter, in which the agent only gets partial, potentially noisy, observations over the true state of the environment. Especially, we aim to answer:
\begin{center} \vspace{-2pt}
    \emph{How do we maximize the entropy of the true states \\ with a policy conditioned on observations only?}
\end{center} \vspace{-2pt}
First, we model this setting through a Partially Observable MDP~\citep[POMDP,][]{aastrom1965optimal} formalism, in which the agent's observations are generated from an observation matrix conditioned on the true state.
We consider two distinct learning settings in which the specification of the POMDP is either known or unknown to the agent, respectively.

The former is motivated by domains in which we can train the agent's policy on a simulator of the environment and then deploy the optimal policy in the real world. %In the rescue operation instance described above, we could address the learning problem in computerized 3D models of the target terrain, before deploying the robot on the actual task. 
Still, the problem is non-trivial, as we have to train a policy taking the input available at deployment. Even solving a known POMDP is an intractable problem~\cite{mundhenk2000complexity} as it requires exponential time in general. Moreover, the problem is optimized by policies conditioned on the history of observations~\citep[e.g.,][]{bertsekas1976dynamic}, which are intractable to store.
We sidestep the \emph{memory} complexity by defining a clever policy class in which the action distribution is conditioned on a function of the current \emph{belief} on the state of the environment, which is a popular model of uncertainty in POMDPs~\cite{kaelbling1998planning}. Then, we overcome the computational complexity by considering a first-order relaxation of the state entropy objective, which we optimize via policy gradient~\cite{williams1992simple, sutton1999policy}, a methodology that has been previously considered for state entropy maximization in MDPs~\citep[e.g.,][]{mutti2021task}.

However, a simulator is not available in all the relevant applications. Can we still learn a reasonable policy in those settings? When the POMDP is unknown, we cannot access the entropy over the true states to compute the gradient, as we only have observations. A naive sidestep is to compute the gradient of the entropy over observations, optimizing an objective function that is called Maximum Observation Entropy (MOE). On a recent study of strengths and limitations of MOE, \citet{zamboni2024rlc} show that the mismatch between the entropy over observations and true states can be significant in relevant domains (e.g., the rescue operation setting we described above).
To overcome this limitation, we can instead compute approximate beliefs from observations~\cite{subramanian2022approximate} and then optimize the entropy of the states sampled from the beliefs as a \emph{proxy} objective that incorporates all of the information available about the entropy on the true states.
We can show that the latter is a better approximation than the trivial entropy of observations in general.

Optimizing the proxy objective still involves a crucial issue: The belief is not completely out of the control of the agent, who has an explicit incentive to take actions that maximize the uncertainty of the belief so that the states sampled from the belief will come with higher entropy. We call the latter the \emph{hallucination} problem, as the agent can hack the objective to make herself/himself believe the entropy on the true states is higher than it actually is. To mitigate this effect, we introduce a \emph{regularization} scheme that penalizes the entropy of the belief so that the agent faces a dueling objective that incentives the entropy of the states sampled from the belief on one side and discourages the agent from pursuing beliefs with higher entropy on the other.
Finally, we design a policy gradient method for the \emph{regularized} objective, for which we provide extensive theoretical and empirical corroborations, showing that the resulting performance nearly matches the one of the policy that maximizes the true objective when good approximators of the belief are available.

\paragraph{Contributions.} We make the following contributions:
\begin{itemize}[topsep=-2pt]
    \item We provide the first generalization of the state entropy maximization problem to POMDPs (Section~\ref{sec:problem_formulation});
    \item We provide a family of tractable policy gradient methods that address first-order relaxations of the introduced problem with known (Section~\ref{sec:known_model}) or unknown (Section~\ref{sec:unknown_model}) POMDP specification, respectively;
    \item We provide extensive theoretical characterizations of the approximation gap and optimization landscape of the introduced objectives (Section~\ref{sec:known_model} and \ref{sec:unknown_model});
    \item We provide an experimental campaign to uphold the design of the introduced algorithms in a variety of illustrative POMDPs (Section~\ref{sec:experiments}).
\end{itemize}

% contributions
% - formulation
% - method(s)
% - theory
% - experiments

\section{Preliminaries}
\label{sec:preliminaries}

In this section, we introduce the notation we will use in the paper and the most relevant background on POMDPs (Section~\ref{sec:preliminaries_pomdps}) and state entropy maximization (Section~\ref{sec:preliminaries_maxent}).

\textbf{Notation.}~~Let $\A$ a set of size $|\A|$. We denote the $T$-times Cartesian product of $\A$ as $\A^T := \times_{t = 1}^T \A$. The simplex on $\A$ is denoted as $\Delta (\A) := \{ p \in [0,1]^{|\A|} \mid \sum_{a \in \A} p(a) = 1 \}$ and $\U (\A)$ denotes a uniform distribution on $\A$. For distributions $p_1, p_2$ we denote $d^\text{TV}(p_1,p_2)$ their total variation distance. We denote as $\mathbb{A} : \A \to \Delta (\B)$ a function from elements of $\A$ to distributions over $\B$. For a function $f : \mathbb{R}^n \to \mathbb{R}$, we denote $\nabla f: \mathbb{R}^n \to \mathbb{R}^n$ its gradient. For vectors $v = (v_1, \dots, v_T)$ and $u = (u_1, \dots, u_T)$, we use $\oplus$ to denote the concatenation $v \oplus u = (v_1, u_1, \dots, v_T, u_T)$.

\subsection{Partially Observable MDPs}
\label{sec:preliminaries_pomdps}

A finite-horizon Partially Observable Markov Decision Process~\citep[POMDP,][]{aastrom1965optimal} is a tuple $\M := (\S, \A, \O, \P, \Obs, T, \mu)$ where $\S$ is a set of states of size $S := |\A|$, $\A$ is a set of actions of size $A := |\A|$, $\O$ is a set of observations of size $O := |\O|$, $\P: \S \times \A \to \Delta (\S)$ is the transition model such that $P(s'|s,a)$ is the probability of reaching $s'$ by taking $a$ in $s$, $\Obs : \S \to \Delta(\O)$ is an observation function such that $\Obs (o | s)$ is the probability of observing $o$ in $s$, $T < \infty$ and $\mu \in \Delta (\S)$ are the horizon and the initial state distribution of an episode, respectively.

In a POMDP, the interaction process goes as follows. At the start of an episode, an initial state is drawn $s_1 \sim \mu$. For each $t < T$, the agent receives an observation $o_t \sim \Obs (\cdot | s_t)$ and plays an action $a_t$, triggering a transition $s_{t + 1} \sim \P (\cdot | s_t, a_t)$. When the final state $s_T$ is reached, the agent observes $o_T \sim \Obs (\cdot | s_T)$ and the episode ends. An episode of interaction returns trajectories over states $\tau_\S = (s_1, \dots, s_T) \in \Tau_\S \subseteq \S^T$, actions $\tau_\A = (a_1, \dots, a_{T-1}) \in \Tau_\A \subseteq \A^{T - 1}$, and observations $\tau_\O = (o_1,\dots, o_{T}) \in \Tau_\O \subseteq \O^T$.

\textbf{Belief.}~~
Crucially, the agent cannot access the true state of the POMDP on which the objective function is usually defined.\footnote{Most of the previous literature in POMDPs define the objective through the maximization of a reward $r: \S \times \A \to \mathbb{R}$. Here, we address a different objective that we will formalize in Section~\ref{sec:problem_formulation}.}
However, it can infer from the observations it receives what is the probability of the process being in a certain state. The latter probability measure, denoted as $\bel \in \B \subseteq \Delta (\S)$ is called a \emph{belief}~\cite{kaelbling1998planning}. The belief is updated following the Bayes rule. The prior is typically set as $\bel_1 = \U (\S)$. Then, for each $1 < t \leq T$, the posterior of the belief having taken action $a_{t - 1} = a$ and received observation $o_t = o$ is computed as 
\begin{equation*}
    \bel_{t}^{ao} (s)
    = \frac{\Obs (o | s) \sum_{s' \in \S} \P (s | s', a) \bel_{t - 1} (s')}{\sum_{s' \in \S}\Obs (o | s') \sum_{s'' \in \S}  \P (s' | s'', a) \bel_{t - 1} (s'')}.
\end{equation*}
In this sense, the elements of $\B$ can be seen as \emph{belief states} evolving according to the belief-update operator $\op^{ao} : \B \to \B$ such that $\bel' = \op^{ao} (\bel)$. In the same way as for states, actions, and observations, an episode of interaction generates a trajectory over beliefs $\tau_{\B} = (\bel_1, \ldots, \bel_T) \in \Tau_\B \subseteq \B^T$.

\textbf{Policies.}~~
We denote the information available to the agent in a given time step as the \emph{information set} $i \in \I$. A policy $\pi : \I \to \Delta (\A) \in \Pi_\I$ describes the action selection strategy of the agent, such that $\pi (a | i)$ denotes the probability of taking $a$ given information $i$ and $\Pi_\I$ is the policy space with information $\I$. We will later specify the meaning of $\I$, which will be either $\O$, $\Tau_\O$, or $\Tau_\B$ according to the setting.\footnote{Introducing the information set allows us to work with stationary Markovian policies on the information, which can be non-Markovian policies w.r.t. states or observations.}
% A policy $\pi$ defines the behavior of an agent interacting with a POMDP, i.e., the strategy according to which an action is selected at any step of the episode. Such strategies are mappings from the information available to the agent with a distribution over actions. A generic (stationary) Markovian policy class can be defined based on an information set $\I$, namely:
% \begin{equation*}
%     \Pi_{\I}: \forall \pi \in \Pi_\I \quad | \quad \pi: \I \to \Delta(\A)
% \end{equation*}

\textbf{Distribution over Trajectories.}~~
Interacting with a POMDP with a fixed policy induces a specific probability distribution over the generated trajectories. Since we have several trajectories generated simultaneously, we denote the joint trajectory as $\tau = \tau_\S \oplus \tau_\A \oplus \tau_\O \oplus \tau_\B$, which probability of being generated under $\pi$ is given by
$
    p^\pi (\tau) =
     \mu (s_1) \prod_{t=1}^T \Obs(o_t|s_t) \pi(a_t | i_t) \P(s_{t+1}|s_t, a_t) \op^{o_t a_t}(\bel_{t+1}|\bel_{t}).
$

%We denote $\bar p_\S (\tau_\S) = \sum_{ \tau_\A, \tau_\O, \tau_\B}p^\pi(\tau)$ the marginal probability of $\tau_\S$ in $\tau$, while the conditional probability is $p^\pi(\cdot|\tau_\S)$.
Interestingly, the belief state formulation allows to extract from $\tau$ \emph{believed trajectories} as well, i.e., trajectories $\tau_{\tilde \S} = (\tilde s_1, \dots, \tilde s_T) \in \Tau_{\tilde \S} \subseteq \S^T$ where the states, called \emph{believed} states, are not the true states of the POMDP but samples from the belief $\tilde s_t \sim \bel_t$, which are generated with probability
$
    p(\tau_{\tilde \S}\vert \tau_\B) = \prod_{t = 1}^T \bel_t(\tilde s_t).
$

\textbf{Distribution over States.}~
A trajectory $\tau_\S$ obtained from an interaction episode induces an empirical distribution over true states $d(\tau_\S) = (d_{s_1}(\tau_\S), \ldots, d_{s_S}(\tau_\S))$ such that $d_{s_i}(\tau_\S) = \frac{1}{T}\sum_{s_t \in \tau_\S} \1{s_t = s_i}$. This concept can be generalized to any finite set as well, leading to distributions over observations, belief states, and believed states.

\subsection{State Entropy Maximization}
\label{sec:preliminaries_maxent}

A POMDP such that $\O = \S$ and $\Obs (s | s) = 1$ reduces to a finite-horizon Markov Decision Process~\citep[MDP,][]{puterman2014markov} $\M := (\S, \A, \P, T, \mu)$. The true state of the system is fully observable in MDPs, which means the agent can take actions according to a policy $\pi: \S \to \Delta (\A)$.

In the absence of a reward to be maximized, \citet{hazan2019provably} proposed a \emph{Maximum State Entropy} (MSE) objective
\begin{equation}
    \label{eq:mdp_entropy}
    \max_{\pi \in \Pi_{\S}} \ \Big\{ H(d^\pi) := - \sum_{s \in \S} d^\pi (s) \log (d^\pi (s)) \Big\}
\end{equation}
where $d^\pi := \EV_{\tau_\S \sim p^\pi} [ d (\tau_\S) ]$ is the expected state distribution and $H(d^\pi)$ its entropy. The latter objective is known to be \emph{non-concave} w.r.t. the policy yet to admit a dual formulation that is concave w.r.t. the state distribution~\cite{hazan2019provably}, which allows for efficient computation of an optimal (stochastic in general) Markovian policy.

\citet{mutti2022importance} later formulated a \emph{single-trajectory} version of the state entropy maximization objective,
\begin{equation}
    \label{eq:mdp_entropy_single_trial}
    \max_{\pi \in \Pi_{\I}} \ \Big\{J^\S(\pi) := \EV_{\tau_\S \sim p^\pi} [ H (d(\tau_\S)) ] \Big\}
\end{equation}
in which the agent seeks to maximize the entropy of the empirical state distribution induced in one single trajectory rather than in multiple trajectories (as in Eq.~\ref{eq:mdp_entropy}). The optimal policy in Eq.~\ref{eq:mdp_entropy_single_trial} is known to be deterministic non-Markovian ($\Pi_\I \subseteq \Pi_{\Tau_\S}$) in general, which makes the optimization problem computationally hard~\cite{mutti2022importance}.
\section{Problem Formulation}
\label{sec:problem_formulation}
%\mm{What's the problem we address?} {\color{cyan} - state entropy objective on true states (finite trial) - learn a policy from belief to actions }
State entropy maximization is particularly challenging in POMDPs as the objective function is defined on a space to which the agent has no direct access. %Reaching an even coverage over such a space remains essential in many scenarios described before.
It is clear that the ideal goal of maximizing the MSE objective in Eq.~\ref{eq:mdp_entropy_single_trial} as in (fully observable) MDPs is far-fetched under these premises. 
Addressing MSE in POMDPs includes the following additional and intertwined challenges: \textbf{(a)} Defining a proxy objective function compatible with the setting, i.e., on quantities the agent can observe; \textbf{(b)} Defining a compact policy class such that policies can be efficiently stored. 

In this paper, we will build upon the single-trajectory formulation of the MSE objective, which is closer to the need for practical applications~\cite{mutti2022importance}. We notice that the common infinite-trajectories relaxation considered in previous works~\citep[e.g.,][]{hazan2019provably} is still intractable in POMDPs, which leaves minimal benefit over the single-trajectory formulation (see~\cref{apx:dual_POMDP} for details).

\textbf{(a) Proxy Objective Functions.}~~Optimizing Eq.~\ref{eq:mdp_entropy_single_trial} is ill-posed in POMDPs without further assumptions because states are not observed. We then seek to design proxy objectives whose maximization leads to policies with good performance on the (ideal) original MSE objective as well. The first and most intuitive choice is to formulate an analogous objective over observations instead of states. The (single-trajectory) \emph{Maximum Observation Entropy} (MOE) is
\begin{equation}\label{eq:pomdp_obs_entropy_single_trajectories}
    \max_{\pi \in \Pi_{\I}} \ \Big\{J^\O(\pi) := \EV_{\tau_\O \sim p^\pi} [ H (d(\tau_\O)) ] \Big\}
\end{equation}
While being rather intuitive, this objective is intrinsically problematic. There can be significant mismatches between observation and state spaces. When the POMDPs are under (respectively over) complete~\cite{liu2022partially}, i.e., when the number of observations is less (respectively more) than the number of states, it may be hard to link entropy over observations to entropy over states. Moreover, even when $\O = \S$, a random emission function $\Obs$ could jeopardize any estimate of the state entropy that is based on the entropy of observations. A formal study of the limitations of MOE has been provided in~\cite{zamboni2024rlc}, which characterizes the settings where the entropy of observation is not enough. To address the latter settings, we introduce more reliable proxy objectives in Section~\ref{sec:known_model}, \ref{sec:unknown_model} along with corresponding assumptions on the information available to the agent.

\textbf{(b) Deployable Policy Classes.}~~So far, we denoted the policy class as $\Pi_\I$ for a generic set $\I$ of the available information. An essential point to be addressed in POMDPs is which policy class to use~\cite{cassandra1998exact}. We say a policy class is \emph{deployable} if its policies are conditioned on the information set $\I$ that is available to the agent \emph{at deployment}.\footnote{Even in the case a simulator is available to optimize the policy, we still want to deploy the policy in unknown partially observable environments in general.} We follow a similar definition of deployable policies as for centralized training and decentralized executions in multi-agent settings~\cite{marlbook}. 
It follows that any policy class over true states is not deployable, and this is the case for deterministic non-Markovian policies as well~\cite{mutti2022importance}. Yet, other policy classes are deployable, e.g., over observations, trajectories of observations, and trajectories of beliefs. Ideally, we want to employ the richer deployable policy class, which is the space of non-Markovian policies over observations (or, equivalently, over beliefs). Unfortunately, a policy in this class cannot be efficiently stored in general, so we will look for restricted classes with more reasonable memory requirements. 

\section{MSE with a POMDP Simulator}
\label{sec:known_model}

%\mm{{\color{cyan} Let assume to have the model for training (but not at deployment). Even in this simplified setting, extracting the optimal policy is computationally and memory intensive, our solutions: - computationally efficiency: policy gradient - memory efficiency: restricted policy class - describe the algorithm - analysis (smoothness, gap true feedback vs observation entropy, convergence rate?)}}
First, we consider a simplified setting where:
\begin{assumption}\label{ass:model_known}
    $\P, \Obs$ are fully known in training.
\end{assumption}
This setting encompasses the best-case scenario, in which a (white-box) simulator of the environment is available and the true state of the POMDP can be accessed. Even in this simplified setting, the problem is non-trivial. First, it does not reduce to the MDP problem, as we need to learn a deployable policy. Secondly, the best deployable policy class is problematic in terms of memory complexity. Finally, as the theory demonstrates~\cite{papadimitriou1987complexity, mundhenk2000complexity}, even solving a known POMDP is computationally intractable. These issues drive the algorithmic choices in the following sense: 
\begin{enumerate}[noitemsep, leftmargin=*, topsep=-2pt]
    \item \textbf{Memory complexity.} The policy class will be restricted to memory-efficient policies, such that the policy parameters are polynomial in the size of $\M$.
    \item \textbf{Computational complexity.} A first-order method will be employed, i.e., policy gradient~\cite{williams1992simple, sutton1999policy}, to overcome computational hardness.
\end{enumerate}   

\textbf{(1)}~~Unfortunately, the size of $\Tau_\O, \Tau_\B$ is exponential in $T$, which means that policies over such spaces would require an exponential number of parameters. This leaves the information sets $\O, \B$ as viable options. Similarly, the set $\B$ of belief states reachable in $T$ steps can be extremely large even in simple POMDPs.\footnote{We can compute $\B$ by means of Algorithm~\ref{alg:belief_set} in Appendix~\ref{apx:belief_set}.} Policy classes that are efficient to store are $\Pi_\O$ and $\Pi_{\tilde \S}$, i.e., the set of Markovian policies over observations or believed states. It is known, however, that non-Markovian policies are needed to optimize the single-trajectory MSE in general~\cite{mutti2022importance}. An option is to consider the belief, which is a function of the trajectory over states and actions, as a succinct representation of the history, and then to employ a careful parametrization of the policy to get memory efficiency. Formally, we introduce the \emph{Belief-Averaged} (BA) policy class as $\bar \Pi_{\mathcal B}  := \{ \pi \in \bar \Pi_{\mathcal B}: \pi_\theta (\cdot| \bel) = \langle \theta, \bel \rangle \} \subseteq \Delta(\mathcal A)$.

\textbf{(2)}~~The optimization problem over the latter policy class will be addressed via first-order methods~\cite{williams1992simple}, in order to overcome computational hardness. Previous works have considered policy gradient for MSE in MDPs~\cite{mutti2021task, liu2021behavior}. Here, we derive a specialized gradient for the POMDP setting.\footnote{The full derivation can be found in Appendix~\ref{apx:gradient_computations}.}
\begin{theorem}[Entropy Policy Gradient in POMDPs]\label{th:gradients_generic}
For a policy $\pi_\theta \in \Pi_\I$ parametrized by $\theta \in \Theta \subseteq \mathbb{R}^{IA}$, we have
\begin{equation}
    \nabla_\theta J^i(\pi_\theta) = \EV_{\tau \sim p^\pi}\Big[\nabla_\theta \log \pi_\theta (\tau)H(d(\tau_i))\Big]
\end{equation}
where $\nabla_\theta \log \pi_\theta (\tau) = \sum_t \nabla_\theta \log \pi_\theta(a_t | i_t)$, $i \in \{\S,\O\}$.
\end{theorem}

\begin{algorithm*}[t!]
\caption{{\color{purple} \textbf{Reg-}}PG for MaxEnt POMDPs} \label{alg:pg_pomdp}
\begin{algorithmic}[1]
    \STATE \textbf{Input}: learning rate $\alpha$, initial parameters $\theta_1$, number of episodes $K$, batch size $N$, information set $\I$, proxy class $j \in \{\S, \O,\tilde \S\}$, \textcolor{purple}{regularization parameter $\rho$}
    \FOR{$k = 1$ \textbf{to} $K$}
        \STATE Sample $N$ trajectories $\{\tau^n_j \sim p^{\pi_{\theta_k}}\}_{n \in [N]}$
        \STATE Compute the feedbacks $\{H(d(\tau^n_j))\}_{n \in [N]}$
        \STATE Compute $\{ \log \pi(\tau^n_j)\}_{n \in [N]}$
        \STATE Perform a gradient step
        $\theta_{k+1}  \leftarrow \theta_k + \frac{\alpha}{N}\sum_n^N \log \pi(\tau^n_j)[H(d(\tau^n_j)) {\color{purple} -\rho \sum_t H(b^{n}_t) }] $
    \ENDFOR
    \STATE \textbf{Output}: the last-iterate policy $\pi{_\theta^K}$
\end{algorithmic}
\end{algorithm*}

\textbf{Algorithmic Architecture.}~~
It can be seen that optimizing for different objectives, the policy gradient differs only on the second term of the product, which we refer to as \emph{feedback}. Thus, we propose a general algorithmic framework, which works for any objective, and mimics the structure of REINFORCE~\cite{williams1992simple}. The pseudocode is reported in Algorithm~\ref{alg:pg_pomdp}.\footnote{Note that the meaning and role of the regularization parameters and corresponding regularization term, color-highlighted in the algorithm, will be clarified in the next section.} The main loop of the algorithm ($\mathbf{2}$-$\mathbf{7}$) is composed of the main steps: $(\mathbf{3})$ sample $N$ trajectories with the current policy, $(\mathbf{4})$ extract the feedbacks coherently to the objective being optimized, $(\mathbf{5})$ compute the log-policy term and $(\mathbf{6})$ perform a gradient ascent step over the parameters space.

%\begin{algorithm}[t]
%\caption{PG for MaxEnt POMDPs} \label{alg:pg_pomdp}
%\begin{algorithmic}[1]
%    \STATE \textbf{Input}: learning step $\alpha$, initial parameters $\theta_1$, episodes $K$, batch size $N$, objective $J$, information set $\I$
%    \FOR{$k = 1$ \textbf{to} $K$}
%        \STATE Sample $N$ trajectories $\{\tau^n \sim p^{\pi_{\theta_k}}\}_{n \in [N]}$
%        \STATE Compute the feedbacks $\{H(d(\tau^n))\}_{n \in [N]}$
%        \STATE Compute $\{ \log \pi(\tau^n) = \sum_t^T\nabla_\theta \log \pi_\theta(a^i_t | i^n_t)\}_{n \in [N]}$
        %\STATE Perform a gradient step \\
        %{$ \quad \ \theta_{k+1}  \leftarrow \theta_k + \alpha \frac{1}{N}\sum_n^N \log \pi(\tau^n)H(d(\tau^n)) $}
%    \ENDFOR
%    \STATE return $\pi{_\theta^K}$
%\end{algorithmic}
%\end{algorithm}

\paragraph{Smoothness of the Optimization Landscape.}
We can prove that the considered objectives are locally smooth, making first-order approaches of the kind described above well-suited for the problem.\footnote{The full derivation of the result is in Appendix~\ref{apx:local_lipschitz}.}
%For the objectives $J^\S, J^\O$ the following holds:
\begin{theorem}[Local Lipschitz Constants]
\label{th:lip_constant}
Let $\pi_1, \pi_2 \in \Pi_\I$, let  $\Tau_{i}(\pi_1,\pi_2) = \{\tau_{i} \in \Tau_{i}: p^{\pi_1}(\tau_{i}) >0 \lor p^{\pi_2}(\tau_{i}) >0\}$ be the set of realizable trajectories over $i \in \{\S, \O\}$, and let $\tau^\star_{i} = \argmax_{\tau \in \Tau_{i}(\pi_1,\pi_2)} H(d(\tau))$. It holds
\begin{equation*}
    \vert J^i(\pi_1) - J^i(\pi_2) \vert \leq T  H(d(\tau^\star_i))d^\text{TV}(\pi_1, \pi_2).
\end{equation*}
\end{theorem} 
A global (but looser) upper bound of the Lipschitz constant can be derived as $T H_{\max}$, where $H_{\max}$ is the maximum entropy that can be obtained over the support. These results provide an interesting insight into how (a bound on) the smoothness constant behaves, as both the objectives defined over true states or observations have Lipschitz constants that are not directly dependent on the policies themselves.
\section{MSE without a POMDP Simulator}
\label{sec:unknown_model}

%\mm{What if we don't have the model for training either? This is the most realistic setting after all...} {\color{cyan} - we have to learn from beliefs: This is all we have! - first problem: we have to compute beliefs. To simplify the presentation, we make an oracle assumption with bounded error approximation (belief approximation out of the scope, see these works..) - We could use policy gradient again, but here the second problem: hallucination. How bad can it be? - solution: maximize a lower bound of the true feedback instead - regularized algorithm }

The Assumption~\ref{ass:model_known} of having access to the POMDP specification is rather restrictive and arguably unreasonable in domains where a (white-box) simulator is not available. To overcome this assumption, we aim to refine the design of our algorithmic solution to work with quantities related to observations only. Luckily, beliefs can still be computed approximately well without access to the POMDP model. Belief approximation techniques have been extensively studied in the literature (e.g., see~\cite{subramanian2022approximate} for a summary). Here, we do not delve into the technicalities of the latter, which are out of the scope of this work, and we instead assume to have access to an \emph{approximated} oracle to compute beliefs. 
\begin{assumption}\label{ass:oracle}
    Let $a \in \A$ and $o \in \O$. Given an approximate belief $\hat \bel_t \in \Delta (\S)$ of the true belief $\bel_t$, an \emph{oracle belief approximator} gives $\hat \bel_{t + 1}$ such that $\| \op^{ao}(\hat\bel_{t}) - \hat \bel_{t + 1} \|_1 \leq \epsilon$.
\end{assumption}
With the latter, we can follow \cref{alg:pg_pomdp} as is, computing approximate beliefs instead of the true beliefs.
Yet, we have to change the feedback as we cannot compute the entropy on the true states. Luckily, the trivial MOE feedback~\eqref{eq:pomdp_obs_entropy_single_trajectories} is \emph{not} the only option we have. We can use the approximate beliefs to reconstruct \emph{believed trajectories} over states and then compute the feedbacks as their entropy.
%\footnote{An approximated belief model will be employed, but any model that outputs distributions over states could be employed.}
We call the latter the \emph{Maximum Believed Entropy} (MBE):
\begin{equation}\label{eq:pomdp_hall_entropy_single_trajectories}
    \max_{\pi \in \Pi_{\I}} \ \Big\{ \tilde J(\pi) := \EV_{\tau_\B \sim p^\pi}\EV_{\tau_{\tilde \S} \sim p(\cdot|\tau_\B)} [ H (d(\tau_{\tilde \S}))] \Big\}
\end{equation}
where the update of the belief in $p^\pi$ is now given by the approximate belief oracle. 
%In the latter, the entropy maximization is intertwined with the problem of computing a good proxy for the state distribution.
Notably, we can extend both Theorem~\ref{th:gradients_generic},~\ref{th:lip_constant} to the MBE objective.\footnote{A full derivation can be found in~\cref{apx:gradient_computations}.}
\begin{theorem}[]\label{th:gradients_MBE}
    For a policy $\pi_\theta \in \Pi_\I$ parametrized by $\theta \in \Theta \subseteq \mathbb{R}^{SA}$, we have
\begin{align}
\nabla_\theta \tilde J(\pi) &= \EV_{\tau_\B \sim p^\pi}\EV_{\tau_{\tilde \S} \sim p(\cdot|\tau_\B)}\Big[\nabla_\theta \log \pi_\theta (\tau_{\tilde \S}) [H(d(\tau_{\tilde \S}))]\Big]
\end{align}
where $\nabla_\theta \log \pi_\theta (\tau_{\tilde \S})$ are defined as in~\ref{th:gradients_generic}.
Additionally, let $\Tau_\B (\pi_1,\pi_2) = \{\tau_\B  \in \Tau_\B : p^{\pi_1}(\tau_\B) >0 \lor p^{\pi_2}(\tau_\B ) >0\}$, $\tau^\star_{\B} = \argmax_{\tau \in \Tau_{\B}(\pi_1,\pi_2)} \EV_{\tau \sim \tau_\B}H(d(\tau))$, and $\bar H(\tau_\B^\star)=\EV_{\tau_{\tilde \S} \sim \tau^\star_\B} H(d(\tau_{\tilde \S}))$, we have
\begin{align}
    \vert \tilde J(\pi_1) - \tilde J(\pi_2) \vert &\leq T \bar H(\tau_\B^\star) d^\text{TV}(\pi_1, \pi_2).
\end{align}
\end{theorem}
Interestingly, compared to the other results in Theorem~\ref{th:lip_constant}, MBE displays an upper bound of the Lipschitz constant that depends on the policies $\pi_1, \pi_2$ directly (through $\tau_\B^\star$). Additionally, $\bar H(\tau_\B^\star)$ consists in the best \emph{expected} believed entropy, which is generally smaller than $H(d(\tau^\star_i)), i \in \{\S,\O\}$ of Theorem~\ref{th:lip_constant}.

\textbf{Objectives Gaps and Hallucinatory Effect.}~~Without~\cref{ass:model_known}, we cannot know the value of the MSE objective anymore. Thus, it is hard to keep track of the mismatch between what the agent expects the (latent) performance to be and what it truly is once it is evaluated on the true states of the environment. However, it is possible to show that the true objective lies in a space explicitly encircled by the proxies. First, we provide the following instrumental definitions:
\begin{definition}
    We define $\Tau_\O(\tau_\S) = \{\tau_\O \in \Tau_\O: H(d(\tau_\O)) \geq H(d(\tau_\S))\}, \Tau(\tau_{\S}) = \{\tau_{\tilde\S}\in \Tau_{\tilde\S}: H(d(\tau_{\tilde\S})) \geq H(d(\tau_\S))\}$ as the set of trajectories over observations and believed states, respectively, for which their entropy is higher than the entropy of a fixed trajectory over true states. We let $\P(\Tau_\O|\tau_\S) = \sum_{\tau_\O \in \Tau_\O(\tau_\S)}p^\pi(\tau_\O|\tau_\S), \P(\Tau|\tau_\B) = \sum_{\tau_{\tilde\S} \in \Tau(\tau_\S)} \tau_\B(\tau_\S)$ the cumulative probability of drawing a trajectory form the above sets and $\bar p_\S(\tau_\S) = \EV_{\tau_\B \sim p^\pi(\cdot|\tau_\S)} \P(\Tau|\tau_\B)$ the expected probability of the believed set. Finally, $J^\O(\pi|\tau_\S) = \EV_{\tau_\O\sim p^\pi(\cdot|\tau_\S)}[H(d(\tau_\O))], \tilde J(\pi|\tau_\S) = \EV_{\tau_\B \sim p^\pi(\cdot|\tau_\S)}\EV_{\tau_{\tilde\S}\sim \tau_\B}[H(d(\tau_{\tilde\S}))]$ the MOE (MBE) objective for a fixed trajectory on the states.
\end{definition}

Then, the following theorem holds:

\begin{theorem}[Proxy Gaps]\label{th:proxy_gaps}
For a fixed policy $\pi \in \Pi_\I$, the MSE objective $J^\S(\pi)$ is bounded by the MOE objective according to
\begin{align}
       J^\S(\pi) \leq &\EV_{\tau_\S \sim \bar p_\S}\Big[\frac{1}{\P(\Tau_\O|\tau_\S)} J^\O(\pi|\tau_\S)\Big] \notag\\
      J^\S(\pi) \geq &\EV_{\tau_\S \sim \bar p_\S}\Big[\frac{1}{1-\P(\Tau_\O|\tau_\S)}J^\O(\pi|\tau_\S)\Big]\notag \\&\qquad\qquad-\EV_{\tau_\S \sim \bar p_\S}\Big[\frac{ \P(\Tau_\O|\tau_\S)}{1-\P(\Tau_\O|\tau_\S)}\Big]\log O \notag
\end{align}
Analogously, $J^\S(\pi)$ is bounded by the MBE objective according to
\begin{align}
    J^\S(\pi) \leq &\EV_{\tau_\S \sim \bar p}\Big[\frac{1}{\bar p_\S(\tau_\S)}\tilde J^\S(\pi|\tau_s)\Big]\notag \\
    J^\S(\pi) \geq &\EV_{\tau_\S \sim \bar p}\Big[ \frac{1}{1-\bar p_\S(\tau_\S)}\tilde J^\S(\pi|\tau_\S)\Big] \notag \\ &\qquad\qquad\qquad- \EV_{\tau_\S \sim \bar p}\Big[\frac{\bar p_\S(\tau_\S)}{1-\bar p_\S(\tau_\S)}\Big]\log S \notag
\end{align}
\end{theorem}

\begin{figure}[t]
\begin{center}
\subfigure[$\bar p = 0.02$]{\includegraphics[width=0.1\textwidth]{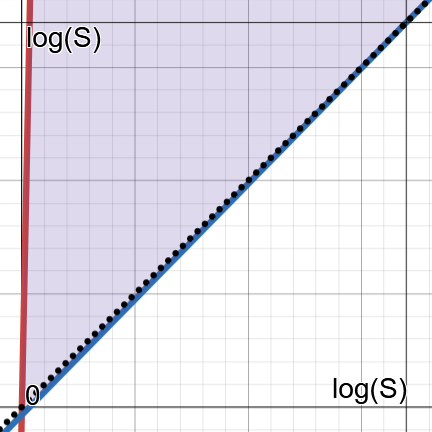}}
\hspace{1em}%
\subfigure[$\bar p = 0.25$]{\includegraphics[width=0.1\textwidth]{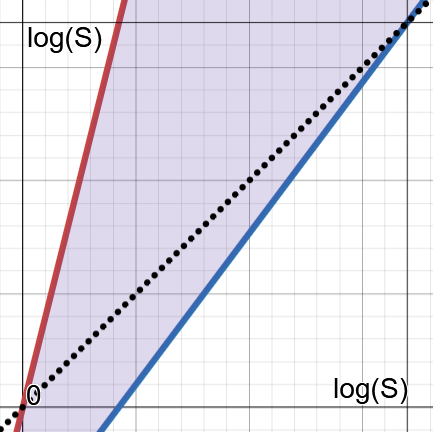}}
\hspace{1em}%
\subfigure[$\bar p = 0.5$]{\includegraphics[width=0.1\textwidth]{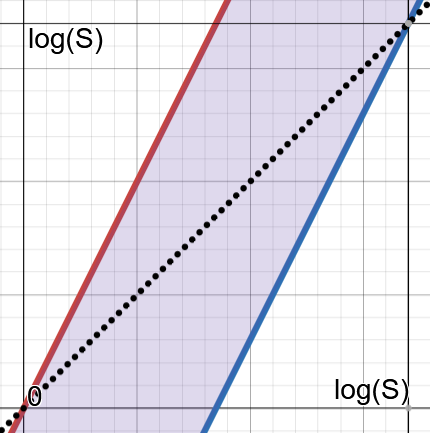}}
\hspace{1em}%
\subfigure[$\bar p = 0.9$]{\includegraphics[width=0.1\textwidth]{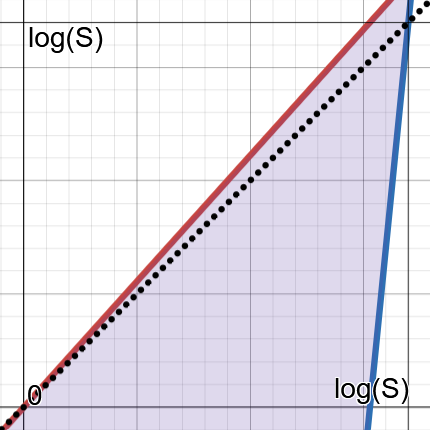}}
\caption{MBE Proxy gaps: for different hallucination probabilities $\bar p_\S$ and a fixed trajectory $\tau_s$, the y-axis represents the \textcolor{RoyalPurple}{\textbf{possible MSE values}} contained between the \textcolor{Purple}{\textbf{upper bound}} and \textcolor{RoyalBlue}{\textbf{lower bound}} as $\mathcal{\tilde J}^\S(\pi|\tau_s)$ varies between $0$ and the maximum value $\log(S)$ (the corresponding \textcolor{black}{\textbf{MBE values}} are plotted over the diagonal to allow a comparisons).\label{fig:MBE_bounds}}
\end{center}
\end{figure}

These results show that the true objective (MSE) is upper/lower bounded by the proxies depending on the probability to generate trajectories (over observations or believed states, respectively) with entropy higher than the one of the trajectory that generated them. We refer to this probability as \emph{hallucination probability} and to the resulting phenomenon as \textbf{hallucinatory effect}. We show in~\cref{fig:MBE_bounds} a visual representation of the MBE gaps in~\cref{th:proxy_gaps}. It is evident that for low over-estimation probabilities $(\bar p_\S = 0.02)$, the MBE objective is a good lower bound for the MSE objective,\footnote{One may notice that the MOE gap is potentially looser: In many scenarios $\log(O) \gg \log(S)$  while on the other hand $\bar p_\S(\tau_\S)$ is the result of an additional expectation with respect to $\P(\Tau_\O|\tau_\S)$.} while it is less so as the hallucination probability increases. The full derivation of these results can be found in~\cref{apx:objectives_gap}.

\begin{figure*}[t!]
    \centering \includegraphics[width=0.4\textwidth]{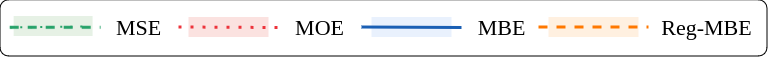}
    \vspace{-0.3cm}
    
    \begin{center}
    \subfigure[(1)][Env. i, deterministic, $0.1$\label{subfig:SingleRoomDeterministicG10}]{\includegraphics[width=4.0cm]{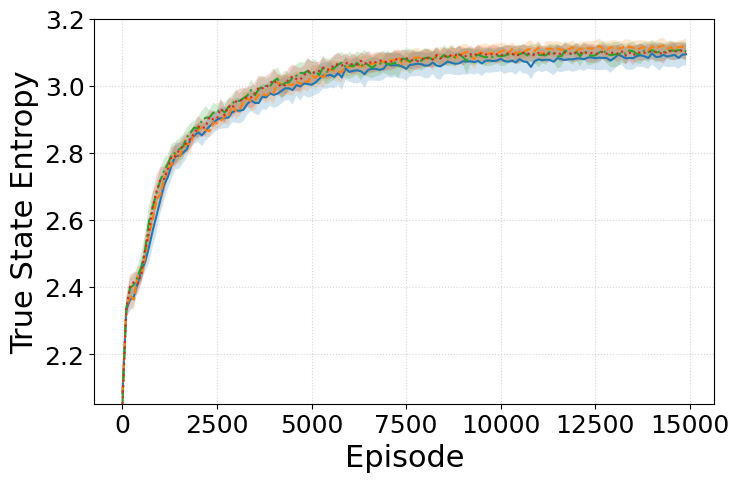}}
    \hfill
    \subfigure[Env. i, deterministic, $10$\label{subfig:SingleRoomDeterministicG01}]{\includegraphics[width=4.0cm]{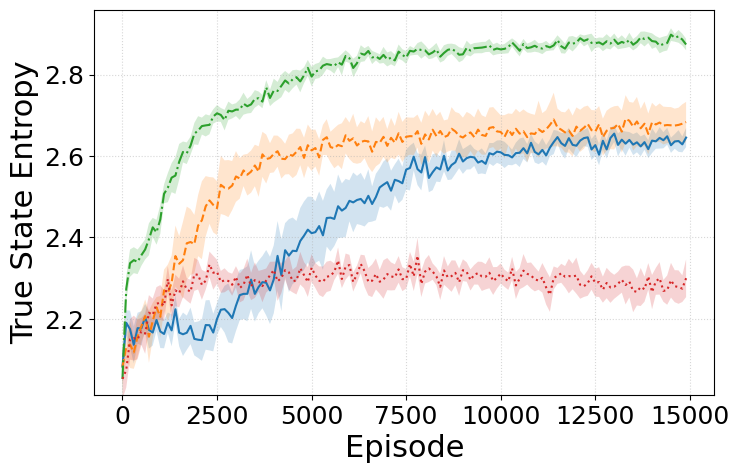}}
    \hfill
    \subfigure[Env. ii, deterministic, $10$ \label{subfig:4RoomsDeterministicG01}]{\includegraphics[width=4.0cm]{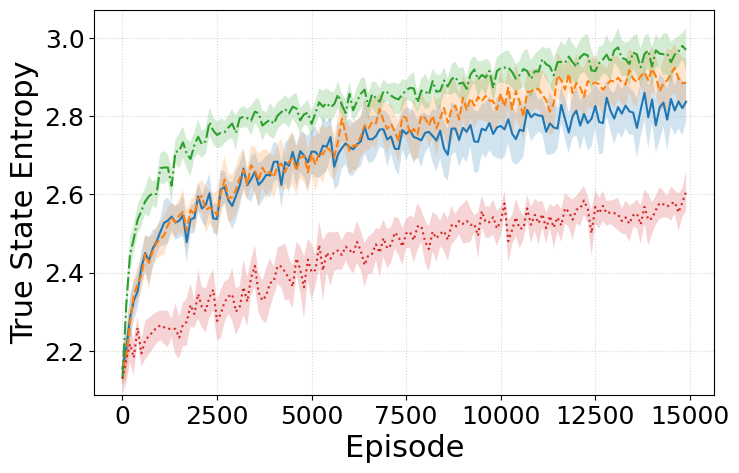}}
    \hfill
    \subfigure[Env. iii, deterministic, n.a. \label{subfig:4ObsDeterministic}]{\includegraphics[width=4.0cm]{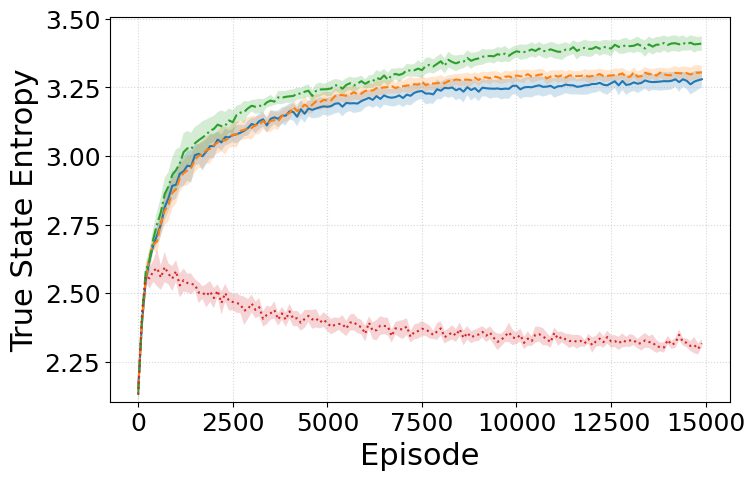}}

    \vspace{-0.2cm}
    \subfigure[Env. i, stochastic, $10$\label{subfig:SingleRoomStochasticG01}]{\includegraphics[width=4.0cm]{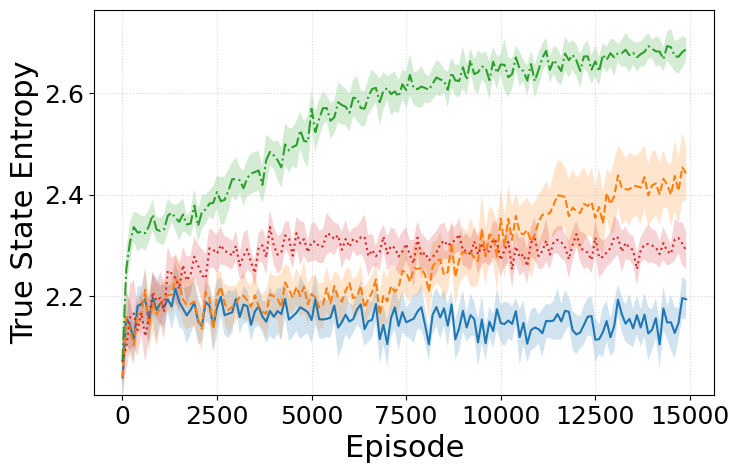}}
    \hfill
    \subfigure[Env. i, stochastic, $10$ \label{subfig:SingleRoomStochasticG01Bounds}]{\includegraphics[width=4.0cm]{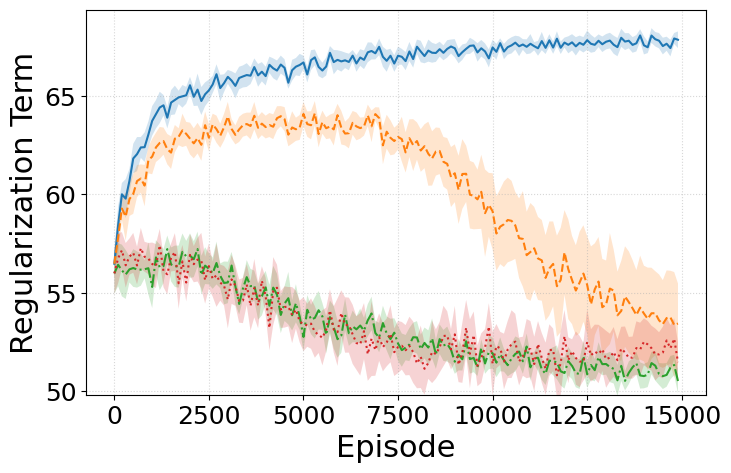}}
    \hfill
    \subfigure[Env. iv, deterministic, n.a.\label{subfig:Deterministic2ObsMultiRoom}]
    {\includegraphics[width=4.0cm, height=2.6cm]{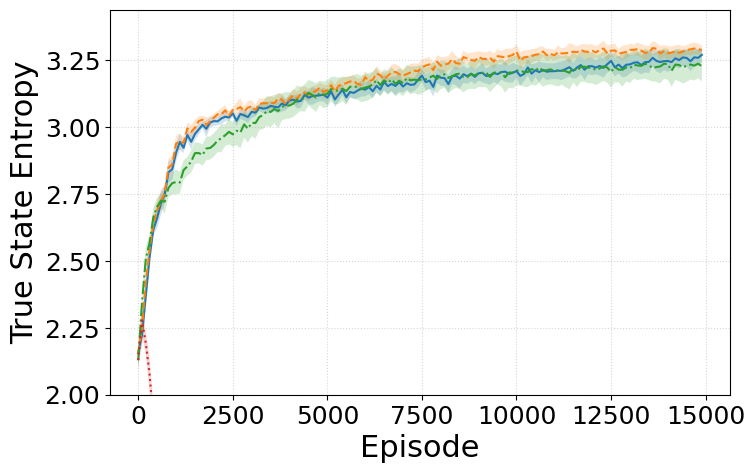}}
    \hfill
    \subfigure[Env. iv, deterministic, n.a.\label{subfig:Deterministic2ObsMultiRoomBound}]
    {\includegraphics[width=4.0cm]{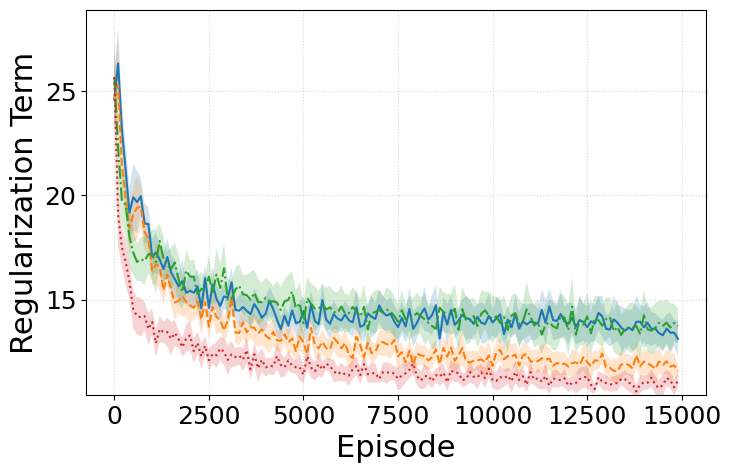}}
    \vspace{-0.3cm}
    \caption{\emph{True state entropy} (or \emph{regularization term}) obtained by Algorithm~\ref{alg:pg_pomdp} specialized for the feedbacks \emph{MSE, MOE, MBE, MBE with belief regularization} (Reg-MBE). For each plot, we report a tuple (environment, transition noise, observation variance) where the latter is \emph{not available} (n.a.) when observations are deterministic. For each curve, we report the average and 95\% c.i. over 16 runs.}
    \label{fig:2}
    % \vspace{-0.3cm}
    \end{center}
\end{figure*}

The role of hallucinatory effects is crucial. Indeed, when the effect of hallucinations is negligible, the proxy objectives are reasonable lower bounds to the true MSE objective, and optimizing them guarantees at least a non-degradation of the MSE objective. The hallucinatory effect, i.e., generating over-entropic trajectories due to the randomness of the generating process, on either observations or beliefs, can be controlled by reducing the randomness of the generating process itself. Unfortunately, under Assumption~\ref{ass:oracle}, we cannot control the observation model as done in \citet{zamboni2024rlc}. However, we have partial control over the trajectory of beliefs that are generated, as they are (approximately) learned and the belief update is conditioned on the taken action. Thus, we can follow the same rationale and derive a regularized objective built upon $\tilde J(\pi)$. In particular, we can maintain a valid lower bound to the MBE objective while enforcing the generation of a sequence of low-entropy belief states $\tau_\B = (\bel_1, \cdots, \bel_T)$ with the following:
\begin{align}
\tilde J(\pi) &\geq \tilde J(\pi) - \rho \EV_{\tau_\B\sim p^\pi}[H(\tau_\B)] \notag \\&\geq \tilde J(\pi) - \rho \EV_{\tau_\B\sim p^\pi}\left[ \sum\nolimits_t H(\bel_t) \right] \notag =: \tilde J_\rho(\pi) 
\end{align}
where the second inequality is due to the sub-additivity of the entropy. We call the obtained $\tilde J_\rho (\pi)$ \emph{MBE with belief regularization} (Reg-MBE for short). Then, the policy gradient $\nabla_\pi \tilde J_\rho(\pi)$ for this objective is
\begin{equation*}
\nabla_\pi \tilde J(\pi) - \rho \EV_{\tau_\B \sim p^\pi}\Big[\nabla_\theta \log \pi_\theta (\tau_{\B}) \sum\nolimits_t H(\bel_t)\Big].
\end{equation*}
It is easy to see that whenever $\tilde J(\pi)$ is a good proxy (i.e., a tight lower bound) of the true MSE objective, then the regularized objective $\tilde J_\rho(\pi)$ will be a reasonable lower bound as well. Most importantly, the regularization term incentives lower-entropy beliefs, which keeps $\tilde J(\pi)$ in a region where it approximates MSE well. From these considerations, a \emph{belief-regularized} version of the Algorithm~\ref{alg:pg_pomdp} is proposed by simply modifying how the gradient step in $(\mathbf{6})$ is computed, as can be seen in the \textcolor{purple}{regularized version} of~\cref{alg:pg_pomdp}.

\section{Numerical Experiments}
\label{sec:experiments}

\begin{figure*}[t!]  \label{fig:oracle_comparison_plotting}
    \begin{center}
        \includegraphics[width=0.9\textwidth]{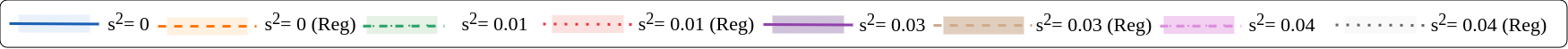}
        \vspace{-0.1cm}
        
        \subfigure[Env. i, deterministic, $10$\label{subfig:PerformancesDeterministicSingleRoomG01Oracle}] {\includegraphics[width=4.0cm]{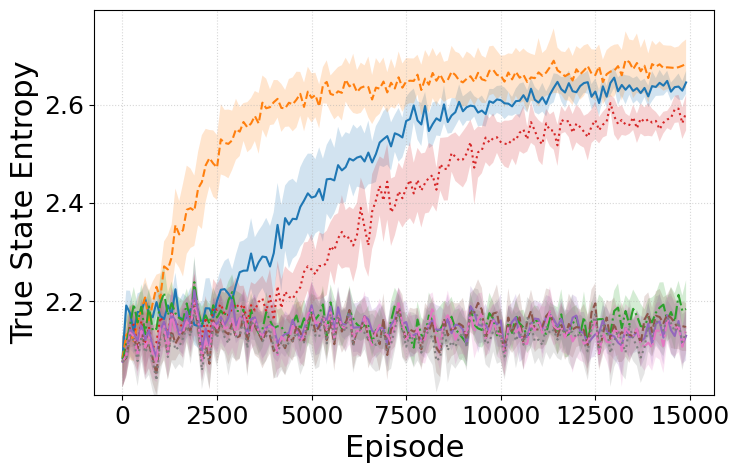}}
        \subfigure[Env. i, stochastic, $10$
        \label{subfig:PerformancesStochasticSingleRoomG01Oracle}] {\includegraphics[width=4.0cm]{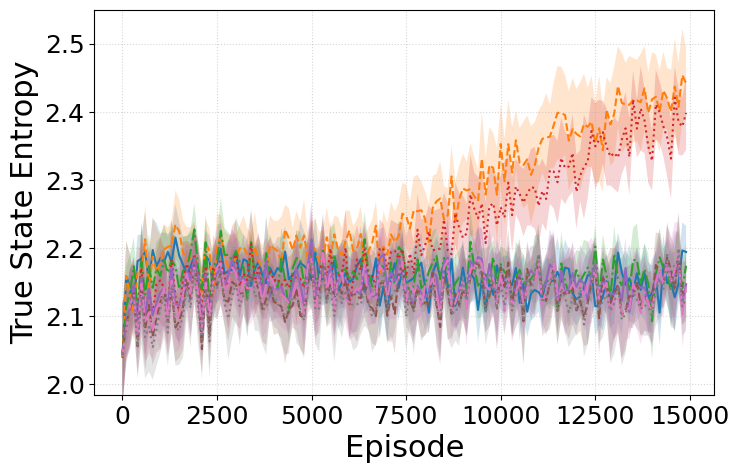}}
        \subfigure[Env. iii , deterministic, n.a.\label{subfig:PerformancesDeterministic4ObsOracle}] {\includegraphics[width=4.0cm]{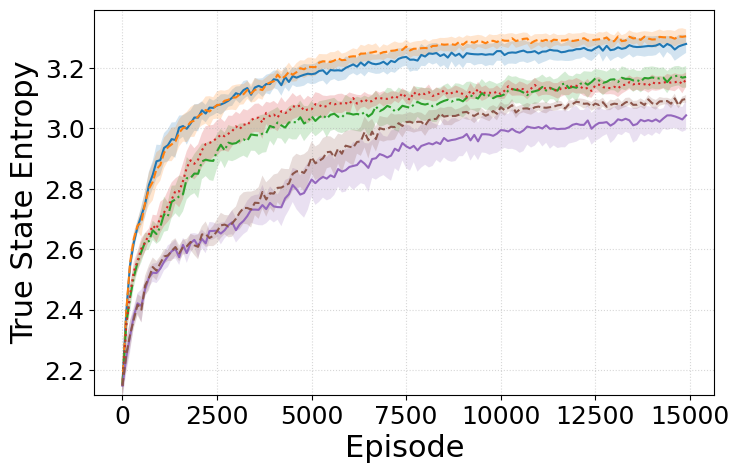}}
        \subfigure[Env. iv, deterministic, n.a.\label{subfig:PerformancesDeterministic2ObsOracle}] {\includegraphics[width=4.0cm]{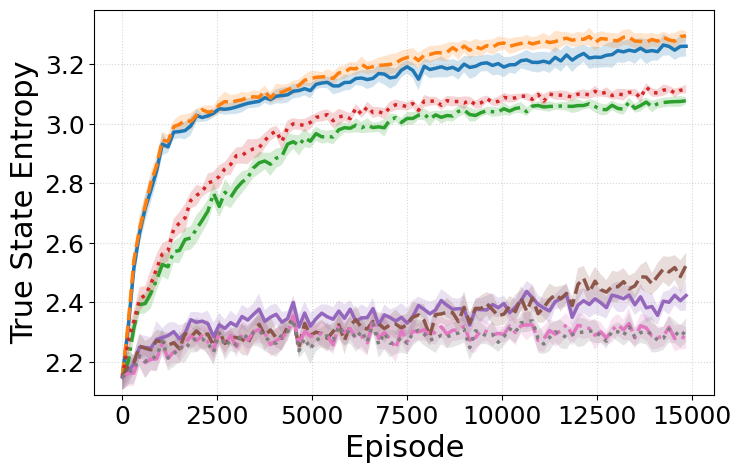}}
        \vspace{-0.3cm}
        \caption{\emph{True state entropy} obtained by Algorithm~\ref{alg:pg_pomdp} with \emph{MBE, MBE with belief regularization} (MBE with Reg) feedbacks under different levels of approximation noise $s^2$. For each plot, we report a tuple (environment, transition noise, observation variance) where the latter is \emph{not available} (n.a.) when observations are deterministic. For each curve, we report the average and 95\% c.i. over 16 runs.}
        \label{fig:3}
    \end{center}
    \vspace{-0.3cm}
\end{figure*}

In this section, we provide an empirical corroboration of the proposed methods and reported claims (results reported in Figure~\ref{fig:2} and \ref{fig:3}). The section is organized as follows: 
\begin{itemize}[noitemsep, topsep=-2pt]
    \item[\ref{subsec:domain_description}] We describe the experimental set-up;
    \item[\ref{subsec:comparison_objectives}] We compare the performance driven by the proxy objectives (MOE, MBE, MBE with belief regularization) against the ideal objective (MSE);
    \item[\ref{subsec:impact_of_approximation}] We study the impact of belief approximation on MBE-based algorithms (with and without regularization).
\end{itemize}

\subsection{Experimental Set-Up}\label{subsec:domain_description}

We consider the following set of finite domains:
\begin{itemize}[noitemsep, topsep=-2pt]
    \item[\textbf{(i)}] A $5\times5$-Gridworld with a single room, where $\O=\S$ and the emission matrix $\Obs$ is such that every row is a (discretized) Gaussian $\Obs(o|s) = \mathcal{N}(s, \sigma^2)$;
    \item[\textbf{(ii)}] A $6 \times 6$-Gridworld with 4 identical rooms, where $\O=\S$ and the emission matrix $\Obs$ is such that every row is a (discretized) Gaussian $\Obs(o|s) = \mathcal{N}(s, \sigma^2)$;
    \item[\textbf{(iii)}] A $6 \times 6$-Gridworld with 4 identical rooms, where $\O = \{1,2,3,4\}$ and the deterministic emission matrix $\Obs$ such that for every state $\Obs (s)$ is the id of the room in which the state lies;
    \item[\textbf{(iv)}]  A $6 \times 6$-Gridworld with 4 identical rooms, where $\O = \{1,2\}$ and the deterministic emission matrix $\Obs$ such that for every state $\Obs (s)$ is the side of the grid (left rooms or right rooms) the state lies in.
\end{itemize}
In all the environments described above, the agent has four actions to take, one for moving to the adjacent grid cell in each of the coordinate directions. Moving against a wall undoes the effect of an action. When we say an environment is \emph{deterministic} we mean that the agent actions never fail. In a \emph{stochastic} environment each action has $0.1$ failure probability, which has the equivalent effect of taking one of the other three actions at random. Finally, we compare the algorithms designed for the MSE, MOE, MBE objectives presented in previous sections.\footnote{While we only compare algorithms of our design, we note that we could not find any previous algorithm addressing state entropy maximization in POMDPs.} Irrespective of the optimized objective, their performance is evaluated on the \textbf{true state entropy} (Equation~\ref{eq:mdp_entropy_single_trial}), which is the ultimate target of state entropy maximization in POMDPs. All of the algorithms optimize a policy within the BA class $\bar \Pi_\B$.
A visualization of the described environment is provided in \cref{apx:environments_visualization}, while the choice of the experimental parameters is discussed in~\cref{apx:parameters}. \cref{apx:policy_class} provides a finer analysis of the choice of the policy class.

\subsection{MSE in POMDP with the Proxy Objectives}\label{subsec:comparison_objectives}

In this section, we compare the performance obtained by Algorithm~\ref{alg:pg_pomdp} specialized for the different proxy objectives. For the sake of clarity, here we assume the belief updates to be computed exactly, while we study the impact of the belief approximation in the next section.

\cref{subfig:SingleRoomDeterministicG10} shows that all of the objectives works equally well in easy settings, e.g., deterministic transitions and small observation noise. However, major differences arise when considering harder settings. 
The MOE objective is sensitive to the quality of the observations, which is evident from the performance degradation in Figures \ref{subfig:SingleRoomDeterministicG01}, \ref{subfig:4RoomsDeterministicG01}, \ref{subfig:4ObsDeterministic}. 
Instead, MBE objectives are remarkably robust to their (diminishing) quality. MBE with belief regularization (Reg-MBE) always performed better than the non-regularized version, showing faster convergence or better final performance.

In Figure~\ref{subfig:SingleRoomStochasticG01}, we see that stochastic transitions are also arduous for MOE and MBE. MBE with belief regularization proved to be better. Interestingly, the true state entropy improvement happens concurrently with the optimization of the regularization term (\cref{subfig:SingleRoomStochasticG01Bounds}). 

Unsurprisingly, optimizing the MSE objective leads to the best performance in most cases, as a testament that whenever the POMDP specification is available in simulation, it is worth training the policy we seek to deploy on the true state entropy. Interestingly, in some limit cases with extreme disentanglement between the observations and the true MSE objective (Figure~\ref{subfig:Deterministic2ObsMultiRoom}), the belief-regularized MBE proxy performed slightly better. 

Finally, Figure~\ref{subfig:SingleRoomStochasticG01Bounds} shows how the MBE is severely hallucinated, an effect that is mitigated with belief regularization.

\subsection{The Impact of Belief Approximation}\label{subsec:impact_of_approximation}

In the previous section, we compared the algorithms in an ideal setting in which the belief is approximated exactly. Here we instead consider the effect of the belief approximation on the same experiments. Especially, to keep full generality of our results, we perturb the exact beliefs with an entry-wise Gaussian noise (with variance $s^2=\{0, 0.01, 0.03, 0.04\}$ respectively), so that our results do not apply to a single belief approximator but any approximator with a bounded error.\footnote{For the sake of clarity, here we report the variance of the perturbation instead of the approximation as in Assumption~\ref{ass:oracle}.}

All Figures from~\ref{subfig:PerformancesDeterministicSingleRoomG01Oracle} to~\ref{subfig:PerformancesDeterministic2ObsOracle} provide two important evidences. First, when good belief approximators are available, the resulting performance is strikingly similar to the ideal setting with exact beliefs. Secondly, MBE with belief regularization is significantly more robust to perturbations, hinting that mitigating hallucination also alleviates the impact of the approximation error to some extent.

\section{Related Work}
\label{sec:related_work}

Below, we summarize the most relevant work on POMDPs, state entropy maximization, and policy optimization.

\textbf{POMDPs.}~~
Learning and planning problems in POMDPs have been extensively studied. In the most general formulation, POMDPs have been shown to be computationally and statistically intractable~\cite{papadimitriou1987complexity, krishnamurthy2016pac}. Nonetheless, several recent works have analyzed tractable sub-classes of POMDPs under convenient structural assumptions~\citep[to name a few][]{jin2020sample, golowich2022planning, chen2022partially, liu2022partially, zhan2022pac, zhong2023gec}. Strides have also been made in modeling beliefs as approximate information states~\cite{subramanian2022approximate} and in the design of practical algorithms~\cite{hafner2019dream}.

\textbf{State Entropy Maximization.}~~
State entropy maximization in MDPs has been introduced in~\citet{hazan2019provably} and then considered in a flurry of subsequent works~\cite{lee2019smm, mutti2020ideal, mutti2021task, mutti2022importance, mutti2022unsupervised, mutti2023unsupervised, zhang2020exploration, guo2021geometric, liu2021behavior, liu2021aps, seo2021state, yarats2021reinforcement, nedergaard2022k, yang2023cem, tiapkin2023fast, jain2023maximum, kim2023accelerating, zisselman2023explore} addressing the problem from various angles.
While~\citet{savas2022entropy} study the problem of maximizing the entropy over trajectories induced in a POMDP, we are the first to formulate \emph{state} entropy maximization in the latter setting.
Complementary results on when the observation entropy is a sensible target for state entropy maximization in POMDPs are reported in the concurrent work~\cite{zamboni2024rlc}.

\textbf{Policy Optimization.}~~
The use of first-order methods~\cite{sutton1999policy, peters2008reinforcement} to address non-concave policy optimization is not new in RL. We considered a \emph{vanilla} policy gradient estimator~\cite{williams1992simple} but several refinements could be made, such as natural gradient~\cite{kakade2001natural}, trust-region schemes~\cite{schulman2015trust}, and importance sampling~\cite{metelli2018policy}.
\section{Conclusion}
\label{sec:conclusion}

In this paper, we generalize the state entropy maximization problem in POMDPs. Especially, we aim to learn a policy that maximizes the entropy over the true states of the environment while accessing partial observations only. In the paper, we show that this entails several critical challenges. First, we propose a family of proxy objectives to approximate the ideal (but not accessible) original objective through quantities that are available to the agent. Then, we choose a convenient sub-class of non-Markovian policies that retain compressed information of history without incurring unreasonable memory requirements. Finally, we designe practical first-order algorithms, which are based on policy gradient, to overcome the inherent non-convexity of the considered objective functions.

Future works can extend our results in many directions, which include integrating a belief approximation method into the algorithmic pipeline~\citep[e.g.,][]{zintgraf2019varibad, subramanian2022approximate}, designing practical implementations for continuous domains~\citep[e.g.,][]{liu2021behavior}, and investigating more policy classes with succinct representations of the history beyond the one we considered.

We believe that our work can be a crucial first step in the direction of extending state entropy maximization to yet more practical settings, in which the state of the system is often not fully observed.

\section*{Impact Statement}
% verbatim from ICLM call for papers
This paper presents work whose goal is to advance the field of Machine Learning. There are many potential societal consequences of our work, none of which we feel must be specifically highlighted here.

%\section*{Acknowledgements}

\bibliography{biblio}
\bibliographystyle{icml2024}

%%%%%%%%%%%%%%%%%%%%%%%%%%%%%%%%%%%%%%%%%%%%%%%%%%%%%%%%%%%%%%%%%%%%%%%%%%%%%%%
%%%%%%%%%%%%%%%%%%%%%%%%%%%%%%%%%%%%%%%%%%%%%%%%%%%%%%%%%%%%%%%%%%%%%%%%%%%%%%%
% APPENDIX
%%%%%%%%%%%%%%%%%%%%%%%%%%%%%%%%%%%%%%%%%%%%%%%%%%%%%%%%%%%%%%%%%%%%%%%%%%%%%%%
%%%%%%%%%%%%%%%%%%%%%%%%%%%%%%%%%%%%%%%%%%%%%%%%%%%%%%%%%%%%%%%%%%%%%%%%%%%%%%%
\newpage
\appendix
\onecolumn
\section{Proofs and Additional Material}
\label{apx:proofs}

\subsection{Infinite-trajectories Formulations}\label{apx:dual_POMDP}

As for MDP theory, the Maximum State Entropy proxies can be formulated in an infinite trajectories form, namely as for Equation \eqref{eq:mdp_entropy} it is possible to write the infinite trajectories formulation of the MOE (MBE) as 
\begin{align*}
    \max_{\pi \in \Pi_{\I}} \ \Big\{ J^\O_\infty = H(d^\pi_\O) := - \sum_{o \in \O} d^\pi_\O (o) \log (d^\pi_\O (o)) \Big\} \\
    \max_{\pi \in \Pi_{\I}} \ \Big\{ \tilde J_\infty = H(d^\pi_\S) := - \sum_{s \in \S} d^\pi_\S (s) \log (d^\pi_\S (s)) \Big\}
\end{align*}
where $d^\pi_\O := \EV_{\tau_\O \sim p^\pi} [ d (\tau_\O) ]$ and $ d^\pi_\S := \EV_{\tau_\B \sim p^\pi}\EV_{\tau_{\tilde \S} \sim p(\cdot|\tau_\B)} [ d (\tau_{\tilde \S}) ]$ is the expected observation (believed state) distribution. These objectives are linked to the single trajectory ones through Jensen's Inequality due to the concavity of the entropy function, namely

\begin{align*}
    J^\O &= \EV_{\tau_\O \sim p^\pi} [ H (d(\tau_\O) ] \leq H(\EV_{\tau_\O \sim p^\pi}d(\tau_\O)) = H( d^\pi_\O) = J^O_\infty \\
    \tilde J &= \EV_{\tau_\B \sim p^\pi}\EV_{\tau_{\tilde \S} \sim p(\cdot|\tau_\B)} [ H (d(\tau_{\tilde \S})] \leq H(\EV_{\tau_\B \sim p^\pi}\EV_{\tau_{\tilde \S} \sim p(\cdot|\tau_\B)} [ d (\tau_{\tilde \S})]) = H(d^\pi_\S) = \tilde J_\infty
\end{align*}

Interestingly, the MBE objective has a clean and neat equivalent formulation in belief-state POMDPs that can be turned into a dual problem as for MDPs, yet the resulting problem is still intractable. More specifically, having defined belief states, we can encode the POMDP $\M$ into a corresponding \emph{belief} MDP $\M_\B := (\B, \A, \wP, \Bf,\bel_0, T)$ where
\begin{itemize}
    \item $\B$ is a finite set of states such that each $\bel \in \B$ corresponds to a belief state, and $\B$ is obtained by running \cref{alg:belief_set} in $\M$;
    \item $\A$ is the set of actions in $\M$;
    \item $\wP: \B \times \A \to \Delta_\B$ is the transition model of the belief MDP defined in a few lines;
    \item $\bel_0 \in \B$ is the initial state;
    \item $T$ is the horizon length.
\end{itemize}
To fully characterize $\M_\B$, we can extract the transition model $\wP$ from $\M$ as
\begin{align*}
    \wP(\bel' | \bel, a) 
    &= \sum_{\{ o \in \O | \bel' = \op^{ao} (\bel)\}} P(o | \bel, a) 
    = \sum_{\{ o \in \O | \bel' = \op^{ao} (\bel)\}} \sum_{s \in \S} P(o | s) P(s | \bel, a) \notag\\
    &= \sum_{\{ o \in \O | \bel' = \op^{ao} (\bel)\}} \sum_{s \in \S} \Obs (o | s) \sum_{s' \in \S} \bel(s') \P(s | s', a).
\end{align*}
Let us denote as $d^\pi \in \Delta_{\S}$ the expected finite-horizon state distribution induced by a policy $\pi \in \Pi_\I$  on the true (unobserved) states. Then, we can define the objective function of our problem as
\begin{equation}
    \label{eq:entropy_objective_apx}
    \max_{\pi \in \Pi_\I} H(d^\pi) = \min_{\pi \in \Pi_\I} \EV_{s \sim d^\pi} [\log d^\pi (s)]
\end{equation} 
Following standard techniques for MDPs~\citep{puterman2014markov}, we can obtain the optimal planning policy for~\eqref{eq:entropy_objective_apx} by solving the dual convex program
\begin{align*}
    &\underset{\substack{\bm{d} \in \Delta_{\S} \\ \{\bm{\omega_t} \in \Delta_{\B \times \A}\}_{t\in[1:T]}} }{\text{maximize}} \;\; H (d) \\
    &\text{subject to} \;\; \sum_{a' \in \A} \omega_{t+1} (\bel', a') =\sum_{\bel \in \B, a \in \A} \omega_{t}(\bel, a) \wP (\bel' | \bel, a) & \forall \bel' \in \B, \; \forall t \in 1\dots T \\
    &\text{\color{white}subject to} \;\; d(s) = \frac{1}{T}\sum_t^T\sum_{\bel \in \B, a \in \A} \omega_t(\bel, a) \bel(s)& \forall (s, a) \in \S \times \A
\end{align*}
and then obtaining the resulting (non-stationary) policy from the solution $\bm{\omega}^*$ as $\pi_t (a | \bel) = \omega^*_t (\bel, a) / \sum_{a' \in \A} \omega^*_t (\bel, a'), \forall (\bel, a) \in \B \times \A$. As one may notice, while this problem has a neat and concise formulation, the dimensionality of the optimization problem does not scale with the dimension of $\M$.

\subsection{Belief Set Computation}\label{apx:belief_set}
The belief states set reachable in a $T$ step interaction with a POMDP can be computed via the following Algorithm

\begin{algorithm}[H]
\caption{$Belief\_set(\bel, \B, t, T)$}
\label{alg:belief_set}
\begin{algorithmic}
    \STATE \textbf{Input}: belief $\bel$, set $\B$, step $t$, horizon $T$
    \IF{$t < T$}
        \FOR{ $(o, a) \in \O \times \A$}
            \STATE $\bel' = T^{ao} (b)$
            \IF{$\bel' \notin \B$}
                \STATE $\B = Belief\_set (\bel', \B \cup \{\bel' \}, t + 1, T)$
            \ENDIF
        \ENDFOR
    \ENDIF
    \STATE return $\B$
\end{algorithmic}
\end{algorithm}

\subsection{Proofs of Theorem \ref{th:gradients_generic} \& Theorem \ref{th:gradients_MBE}: Policy Gradients Computation}\label{apx:gradient_computations}

Let us denote $\tau = \tau_\S \oplus \tau_\O \oplus \tau_\B\oplus \tau_{\tilde \S}$ and for a generic $i \in \{\S,\O,\B,\tilde\S\}$ we denote $\tau|_i$ as the trajectory $\tau_i$ extracted from $\tau$. This is done to be able to use any kind of policy class considered in the main paper as well. For a generic single trajectory objective defined with $J \in \{J^\S,J^\O, \tilde J\}$ it is possible to write:

\begin{align*}
    \nabla_\theta J(\pi) &=  \nabla_\theta \EV_{\tau \sim p^\pi}[H(d(\tau|_I))]
    \\
    &= \nabla_\theta \sum_{\tau}p^\pi(\tau) H(d(\tau|_i)) \\
    &=  \sum_{\tau} \Big( \nabla_\theta p^\pi(\tau)\Big) H(d(\tau|_i)) \\ 
    \intertext{Thanks to the usual log-trick}
    &=  \sum_{\tau} p^\pi(\tau) \Big( \nabla_\theta \log p^\pi(\tau)\Big) H(d(\tau|_i)) \\
    &=  \EV_{\tau \sim p^\pi}\big[\nabla_\theta \log p^\pi(\tau) H(d(\tau|_i))\big]
\end{align*}

The computation of the gradient is then reconducted to the calculation of the log-policy term $\nabla_\theta \log p^\pi(\tau)$ for the generic class $\pi \in \Pi_\I$. It follows that

\begin{align*}
    \nabla_\theta \log p^\pi(\tau) &= \nabla_\theta \log \Big(\mu (s_1) \prod_{t=1}^T \O(o_t|s_t) \pi(a_t | i_t) \P(s_{t+1}|s_t, a_t) \op^{o_t a_t}(b_{t+1}|b_{t})\Big) \\
    &=\nabla_\theta \Big(\log(\mu (s_1)) + \sum_{t=1}^T\log( \O(o_t|s_t)) + \log (\pi(a_t | i_t)) + \log(\P(s_{t+1}|s_t, a_t)) + \log(\op^{o_t a_t}(b_{t+1}|b_{t}))\Big)
\end{align*}
Where the only terms depending on $\theta$ are indeed the $\I$-specific log-policy terms, leading to
\begin{align*}
    \nabla_\theta \log p^\pi(\tau) &= \sum_t^T\nabla_\theta \log \pi_\theta(a_t | i_t)
\end{align*}
which leads to the standard REINFORCE-like formulation of policy gradients.

\subsection{Proofs of Theorem \ref{th:lip_constant} \& Theorem \ref{th:gradients_MBE}: Lipschitz Constants Computation}\label{apx:local_lipschitz}
\paragraph{MSE/MOE (Theorem \ref{th:lip_constant}):}
Let us define the set of reachable trajectories in $T$ steps by following a generic policy $\pi_i$ as $T_i = \{\tau \in T_i : p^{\pi_i}(\tau)>0\}$, it follows that for both MSE and MOE objective, by defining $\tau$ as $\tau_\S$ or $\tau_\O$ respectively, we can see that
\begin{align*}
\vert J(\pi_1) - J(\pi_2) \vert &= \Big\vert \EV_{\tau \sim p^{\pi_1}} [H(d(\tau))] - \EV_{\tau\sim p^{\pi_2}}  [H(d(\tau))] \Big\vert \\
&\leq  \sum_{\tau\in T_1 \cup T_2} H(d(\tau)) \Big\vert p^{\pi_1}(\tau) - p^{\pi_2}(\tau)\Big\vert \\
\intertext{By defining $\tau^\star \in \argmax_{\tau\in T_1 \cup T_2}H[d(\tau)]$}
&\leq H[d(\tau^\star)]\sum_{\tau \in T_1 \cup T_2}\Big\vert p^{\pi_1}(\tau) - p^{\pi_2}(\tau)\Big\vert \\
\intertext{We notice that $p^{\pi_i} = \prod^\pi_t p^{\pi_i}_t$ and that the total variation between two product distributions can be upper-bounder by the summation over the per-step total variations, namely $d^\text{TV}(\prod^\pi_t p^{\pi_i}_t,\prod^\pi_t p^{\pi_j}_t) \leq \sum_t^T d^\text{TV}(p^{\pi_i}_t,p^{\pi_j}_t)$, leading to}
&=  H[d(\tau^\star)]d^\text{TV}(p^{\pi_1}, p^{\pi_2}) \\
&\leq   H[d(\tau^\star)] \sum_t^T d^\text{TV}( p^{\pi_1}_t, p^{\pi_2}_t)\\
\intertext{The only difference between the two distributions (for a fixed step) consists of the policies}
&= T  H[d(\tau^\star)] d^\text{TV}(\pi^1, \pi^2) = \mathcal{L}(\pi_1,\pi_2)d^\text{TV}(\pi^1, \pi^2)
\end{align*}
It follows a (bound on a) Lipschitz constant dependent on the two policies to be compared that is directly proportional to the best single trajectory (in terms of entropy) reachable by the policies themselves. Any policy able to generate a maximum entropic trajectory will have the highest possible Lipschitz constant. The constant then gets steeper as the quality of the policies improves.
\paragraph{MBE (Theorem \ref{th:gradients_MBE}):}Similarly to the previous steps, 
\begin{align*}
\vert \tilde J(\pi_1) - \tilde J(\pi_2) \vert &= \Big\vert \EV_{\tau_\B \sim p^{\pi_1}} \EV_{\tau_{\tilde \S} \sim \tau_\B(\cdot)} [H(d(\tau_{\tilde \S}))]- \EV_{\tau_\B\sim p^{\pi_2}} \EV_{\tau_{\tilde \S} \sim \tau_\B(\cdot)} [H(d(\tau_{\tilde \S}))] \Big\vert \\
&\leq  \sum_{\tau_\B\in T_1 \cup T_2} \sum_{\tau_{\tilde \S}}\tau_\B(\tau_{\tilde \S}) H(d(\tau_{\tilde \S})) \Big\vert p^{\pi_1}(\tau_\B) - p^{\pi_2}(\tau_\B)\Big\vert \\
&=  \sum_{\tau_\B\in T_1 \cup T_2} \EV_{\tau_{\tilde \S} \sim \tau_\B} H(d(\tau_{\tilde \S})) \Big\vert p^{\pi_1}(\tau_\B) - p^{\pi_2}(\tau_\B)\Big\vert \\
\intertext{Again let us define $\tau^\star_{\B} \in \argmax_{\tau_\B\in T_1 \cup T_2}\EV_{\tau_{\tilde \S} \sim \tau_\B} H(d(\tau_{\tilde \S}))$}
&\leq \EV_{\tau_{\tilde \S} \sim \tau^\star_\B} H(d(\tau_{\tilde \S}))d^\text{TV}( p^{\pi_1},p^{\pi_2})\\
&\leq \EV_{\tau_{\tilde \S} \sim \tau^\star_\B} H(d(\tau_{\tilde \S}))\sum_t^Td^\text{TV}( p^{\pi_1}_t, p^{\pi_2}_t) \\
&= T\EV_{\tau_{\tilde \S} \sim \tau^\star_\B} H(d(\tau_{\tilde \S}))d^\text{TV}(\pi^1, \pi^2) = \mathcal{ \tilde  L}(\pi_1,\pi_2)d^\text{TV}(\pi^1, \pi^2)
\end{align*}
Again, the (local) Lipschitz constant $\mathcal{ \tilde L}(\pi_1,\pi_2)$ is dependent on the maximum (expected) entropy that can be induced by one of the policies.
One may notice that $ \mathcal{L}(\pi_1,\pi_2)$ will be usually higher than $ \mathcal {\tilde L}(\pi_1,\pi_2)$.

\subsection{Proofs of Theorem \ref{th:proxy_gaps}: Proxy Gaps}\label{apx:objectives_gap}

\paragraph{MOE:} Let us define the set of observation-trajectories that have an entropy higher than the entropy of a fixed trajectory over true states, namely $\Tau_\O(\tau_\S) = \{\tau_\O \in \Tau_\O: H(d(\tau_\O)) \geq H(d(\tau_\S))\}$. It follows that by employing the conditional trajectory probability $p^\pi(\tau_\O|\tau_\S)$ one can define the probability $\P(\Tau_\O|\tau_\S) = \sum_{\tau_\O \in \Tau_\O(\tau_\S)}p^\pi(\tau_\O|\tau_\S)$. It follows that 
\begin{align*}
J^\S - J^\O &= \EV_{\tau_\S \sim p^\pi(\cdot)}\Big[H(d(\tau_\S)) -  J^\O(\pi|\tau_\S)\Big]\\
    &=\EV_{\tau_\S \sim p^\pi(\cdot)}\Big[H(d(\tau_\S)) - \EV_{\tau_o\sim p^\pi(\cdot|\tau_\S)}H(d(\tau_o))\Big] \\
    &=\EV_{\tau_\S \sim p^\pi(\cdot)}\Big[H(d(\tau_\S)) - \sum_{\tau_o}p^\pi(\tau_o|\tau_\S)H(d(\tau_o)) \Big]\\
    \intertext{ By definition $H(d(\tau_\O\in \Tau_\O(\tau_\S))) \geq H(d(\tau_\S))$, and by positivity of the entropy function $  H(d(\tau_\O\notin \Tau_\O(\tau_\S))) \geq 0$} 
   &\leq\EV_{\tau_\S \sim p^\pi(\cdot)}\Big[H(d(\tau_\S)) - \P(\Tau_\O|\tau_\S)H(d(\tau_\S))\Big]\\
   &\leq \EV_{\tau_\S \sim p^\pi(\cdot)}\Big[(1-\P(\Tau_\O|\tau_\S))H(d(\tau_\S))\Big] \\
\intertext{It follows that}
       J^\S(\pi) &\leq \EV_{\tau_\S \sim p^\pi(\cdot)}\Big[\frac{1}{\P(\Tau_\O|\tau_\S)} J^\O(\pi|\tau_\S)\Big]
\end{align*}
In the same way, focusing on the terms inside the outer expectation for simplicity, one obtains:

\begin{align*}
J^\S - J^\O &= \EV_{\tau_\S \sim p^\pi(\cdot)}\Big[H(d(\tau_\S)) -  J^\O(\pi|\tau_\S)\Big]\\
    &=\EV_{\tau_\S \sim p^\pi(\cdot)}\Big[H(d(\tau_\S)) - \EV_{\tau_o\sim p^\pi(\cdot|\tau_\S)}H(d(\tau_o))\Big] \\
    &=\EV_{\tau_\S \sim p^\pi(\cdot)}\Big[H(d(\tau_\S)) - \sum_{\tau_o}p^\pi(\tau_o|\tau_\S)H(d(\tau_o)) \Big]\\
    \intertext{Again, one notices that $H(d(\tau_\O \in \Tau_\O(\tau_\S)))) \leq H(d(\tau_\S))$ and $H(d(\tau_\O \in \Tau_\O(\tau_\S)))) \leq \log(O)$, from which the inner expectation turns out to be bounded by the use of the complementary probability $\P(\Tau_\O^C|\tau_\S)= \sum_{\tau_\O \notin \Tau_\O(\tau_\S)}p^\pi(\tau_\O|\tau_\S)$}
    &\geq \EV_{\tau_\S \sim p^\pi(\cdot)}\Big[ (1-\P(\Tau_\O^C|\tau_\S))H(d(\tau_\S)) -  \P(\Tau_\O|\tau_\S)\log(O)\Big] \\
    &= \EV_{\tau_\S \sim p^\pi(\cdot)}\Big[ \P(\Tau_\O|\tau_\S)H(d(\tau_\S)) -  \P(\Tau_\O|\tau_\S)\log(O)\Big] 
\end{align*}

Leading to
\begin{align*}
      J^\S(\pi) \geq  \EV_{\tau_\S \sim p^\pi(\cdot)}\Big[\frac{J^\O(\pi|\tau_\S)-\P(\Tau_\O|\tau_\S)\log (O)}{1-\P(\Tau_\O|\tau_\S)}\Big]
\end{align*}

\paragraph{MBE:} Let us define the similar set for hallucinated trajectories $\Tau(\tau_\S) = \{\tau_{\tilde \S} \in \Tau_{\tilde\S}: H(d(\tau_{\tilde \S})) \geq H(d(\tau_\S))\},  \P(\Tau|\tau_\B) = \sum_{\tau_\S \in \Tau(\tau_\S)} \tau_\B(\tau_\S)$.

\begin{align*}
J^\S(\pi) - \tilde J(\pi) &= 
    \EV_{\tau_\S \sim p^\pi(\cdot)}\Big[H(d(\tau_\S)) - \tilde J^\S(\pi|\tau_\S) \Big]\\ &=     \EV_{\tau_\S \sim p^\pi(\cdot)}\Big[H(d(\tau_\S)) -  \EV_{\tau_\O \tau_\A, \tau_\B\sim p^\pi(\cdot|\tau_\S)}\EV_{\tau_{\tilde \S} \sim \tau_\B}H(d(\tau_{\tilde \S}))\Big]  \\
    &= \EV_{\tau_\S \sim p^\pi(\cdot)}\Big[H(d(\tau_\S)) - \EV_{\tau_\O \tau_\A, \tau_\B\sim p(\cdot|\tau_\S)}\sum_{\tau_{\tilde \S} } \tau_\B(\tau_{\tilde \S})H(d(\tau_{\tilde \S})) \Big]\\
    \intertext{Again $H(d(\tau \in \Tau_\S(\tau_\S))) \geq H(d(\tau_\S))$ and $H(d(\tau \notin \Tau_\S(\tau_\S))) \geq 0$}
   &\leq\EV_{\tau_\S \sim p^\pi(\cdot)}\Big[H(d(\tau_\S)) - \EV_{\tau_\O \tau_\A, \tau_\B\sim p(\cdot|\tau_\S)}[ \P(\Tau|\tau_\B)]H(d(\tau_\S)) \Big]\\
   &\leq \EV_{\tau_\S \sim p^\pi(\cdot)}\Big[(1-\EV_{\tau_\O \tau_\A, \tau_\B\sim p(\cdot|\tau_\S)}
   \P(\Tau|\tau_\B))H(d(\tau_\S))\Big] \\
\intertext{We call $\bar p_\S(\tau_\S) = \EV_{\tau_\B \sim p^\pi(\cdot|\tau_\S)} \P(\Tau|\tau_\B)$, it follows that}
       J^\S(\pi) &\leq \EV_{\tau_\S \sim p^\pi(\cdot)}\Big[\frac{1}{\bar p(\tau_\S)}\tilde J^\S(\pi|\tau_\S)\Big] \\
\intertext{In the same way as before, by simply changing the definitions accordingly, one obtains that:}
      J^\S(\pi) &\geq  \EV_{\tau_\S \sim p^\pi(\cdot)}\Big[\frac{\tilde J^\S(\pi|\tau_\S)-\bar p(\tau_\S)\log S}{1-\bar p(\tau_\S)}\Big]
\end{align*}
\newpage

\section{Experimental Details}
\label{apx:experiments}

\subsection{Code Repository and Reproducibility} 

The code is available at this \href{https://anonymous.4open.science/r/exploration-pomdp-13C9}{link}.

\subsection{Wall-Clock Time for the Main Experiments}

The computational cost of Algorithm \ref{alg:pg_pomdp} is mostly due to the sampling of trajectories in line 3, whereas line 4 and 5 can be computed concurrently to the sampling process and the parameters updates at line 6 are done in parallel. It follows that the computational complexity of Algorithm 1 is
\begin{itemize}
    \item $\mathcal{O} (KT)$ when we can access to parallel simulators for the POMDP;
    \item $\mathcal{O} (NKT)$ when the $N$ trajectories are sampled sequentially (e.g., interacting with a real-world system).
\end{itemize}

Additionally, here we report a table with the wall-clock time for running the main experiments in the paper. Note that all of the experiments take less than an hour of training on general-purpose CPUs.

    \begin{table}[H]
        \centering
        \begin{tabular}{c|c|c|c|c}
        \hline
        Experiment & MSE& MOE & MBE & MBE-Reg\\
        \hline
            Figure 2.a & 1802 & 1797 & 2497 & 2342\\
            Figure 2.b & 1800 & 1791 & 2537 & 2465\\
            Figure 2.c & 2594 &  2603 & 3483 & 3455\\
            Figure 2.d & 2535 &  2510 & 3535 & 3305\\
            Figure 2.e & 1780 & 1803 & 2423 & 2582 \\
            Figure 2.g & 2810 & 2746 & 3452 & 3515\\
        \end{tabular}\caption{Wall-clock time [sec] of the main experiments on general-purpose CPUs.}
    \end{table}

\subsection{POMDP Domains Visualization} \label{apx:environments_visualization}

\begin{figure*}[ht]
\begin{center}
\subfigure[Single Room]{\includegraphics[width=0.2\textwidth]{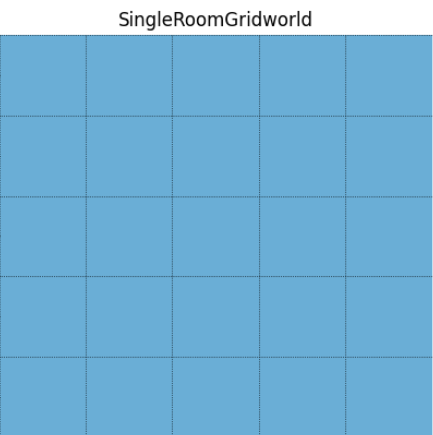}}
\hspace{1em}%
\subfigure[Four Rooms]{\includegraphics[width=0.2\textwidth]{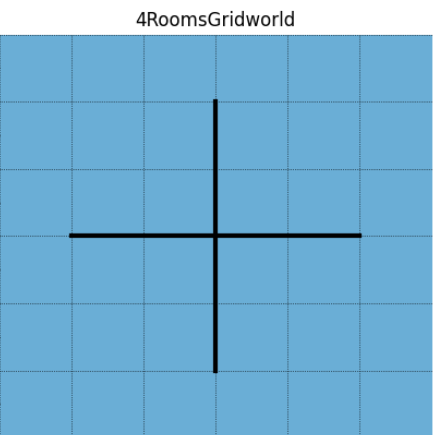}}
\hspace{1em}%
\subfigure[Four Rooms with 4 Observations]{\includegraphics[width=0.2\textwidth]{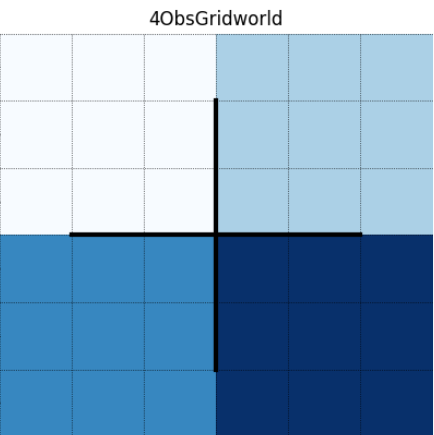}}
\hspace{1em}%
\subfigure[Four Rooms with 2 Observations]{\includegraphics[width=0.2\textwidth]{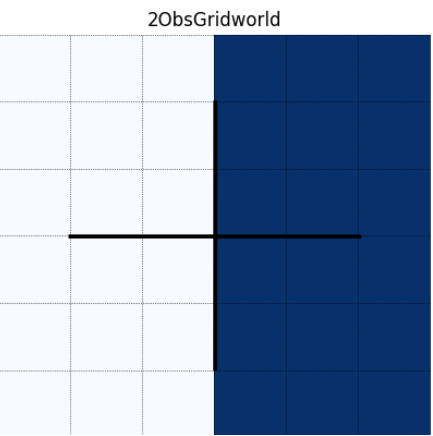}}
\caption{Environments Visualization.\label{fig:environments}}
\end{center}
\end{figure*}

In~\cref{fig:environments} we report a visualization of the four types of domain taken into account. 

\subsection{Choice of Hyperparameters}\label{apx:parameters}

The \textbf{learning rate} was selected as $\alpha=0.3$. The \textbf{batch size} was selected to be $N=10$ after tuning. As for the \textbf{time horizon}, $T = S$ in all the experiments. This makes the exploration task more challenging as every state can be visited at most once. The best regularization term $\rho$ was found to be approximately equal to $0.02$.
%, as can be seen in Fig. \ref{fig:rho}.
\newpage
\subsection{Policy Class Investigation}\label{apx:policy_class}

\begin{figure*}[h]
    \centering \includegraphics[width=0.9\textwidth]{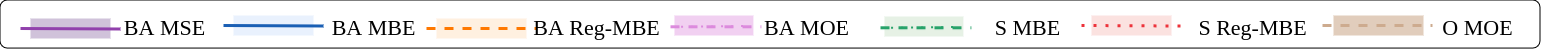}
    \vspace{-0.1cm}
    
    \centering
    \subfigure[(1)][Env. (a), deterministic, $0.1$\label{subfig:apx_SingleRoomDeterministicG10}]{\includegraphics[width=5.0cm]{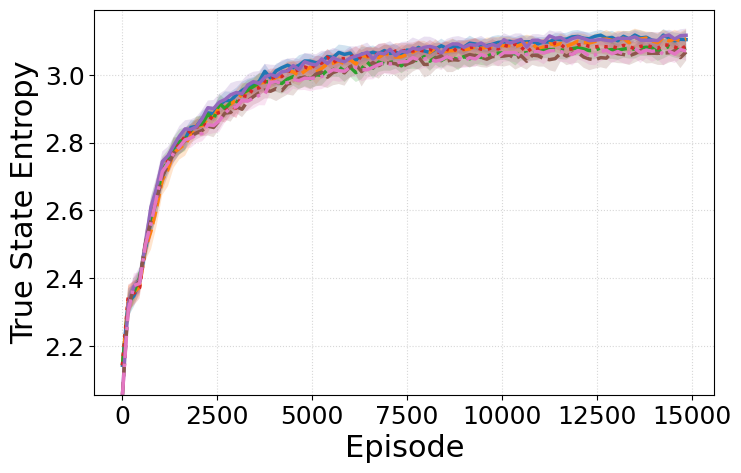}}
    \hfill
    \subfigure[Env. (a), deterministic, $10$\label{subfig:apx_SingleRoomDeterministicG01}]{\includegraphics[width=5.0cm]{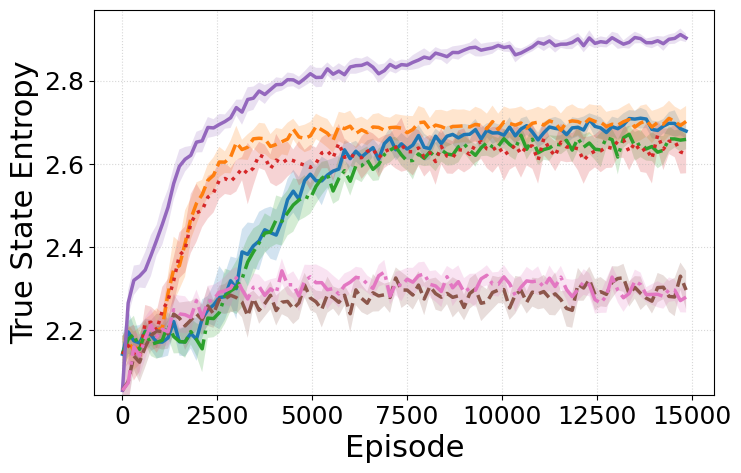}}
    \hfill
    \subfigure[Env. (b), deterministic, $10$ \label{subfig:apx_4RoomsDeterministicG01}]{\includegraphics[width=5.0cm]{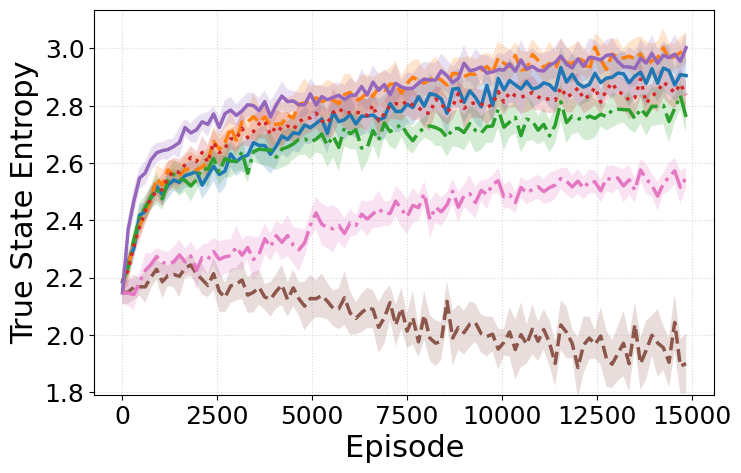}}
    \vspace{-0.2cm}
    \subfigure[Env. (c), deterministic, n.a. \label{subfig:apx_4ObsDeterministic}]{\includegraphics[width=5.0cm]{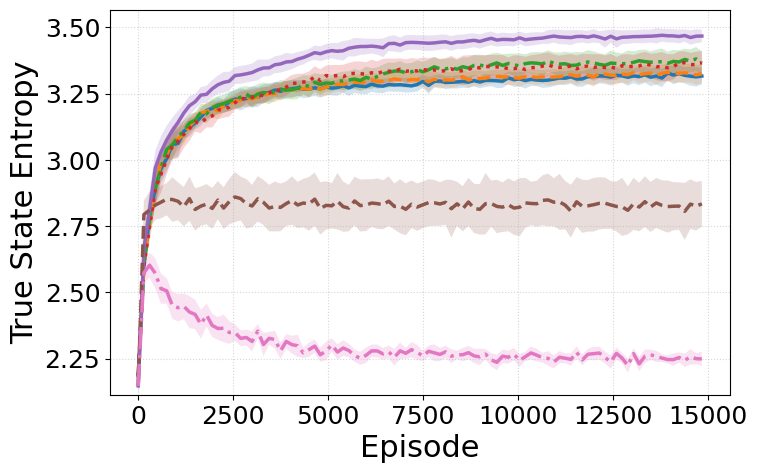}}
    \hfill
    \subfigure[Env. (a), stochastic, $10$\label{subfig:apx_SingleRoomStochasticG01}]{\includegraphics[width=5.0cm]{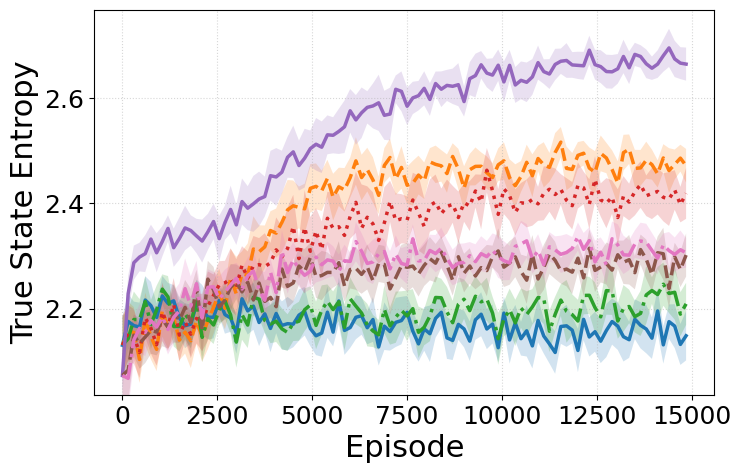}}
    \hfill
    \subfigure[Env. (d), deterministic, n.a.\label{subfig:apx_Deterministic2ObsMultiRoom}]
    {\includegraphics[width=5.0cm]{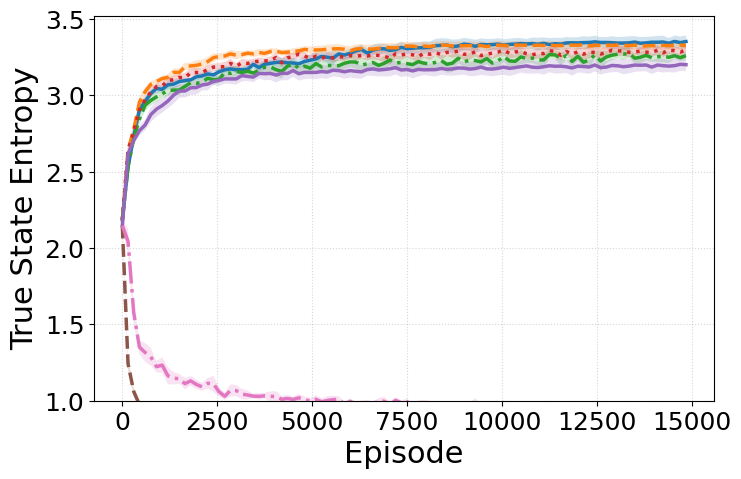}}
    \caption{\emph{True state entropy} obtained by Algorithm~$1$ specialized for the feedbacks \emph{MSE, MOE, MBE, MBE with belief regularization} (Reg-MBE) over different policy classes with direct parametrization: Markovian over observation (O), Belief Averaged (BA), Markovian over hallucinated states (S). For each plot, we report a tuple (environment, transition noise, observation variance) where the latter is \emph{not available} (n.a.) when observations are deterministic. For each curve, we report the average and 95\% c.i. over 16 runs. BA confirms to be the policy class with generally higher performance in all the considered instances.\label{fig:policy_comparison}}
    \vspace{-0.3cm}
\end{figure*}

\begin{figure*}[h!]
    \begin{center}
    \includegraphics[width=0.48\textwidth]{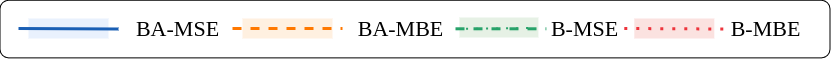} 
    
        %\hfill
        \subfigure[Env. (a) $(|\mathcal S| = 9)$, deterministic, $0.2$\label{subfig:dynamic3x3}]{\includegraphics[width=0.33\textwidth]{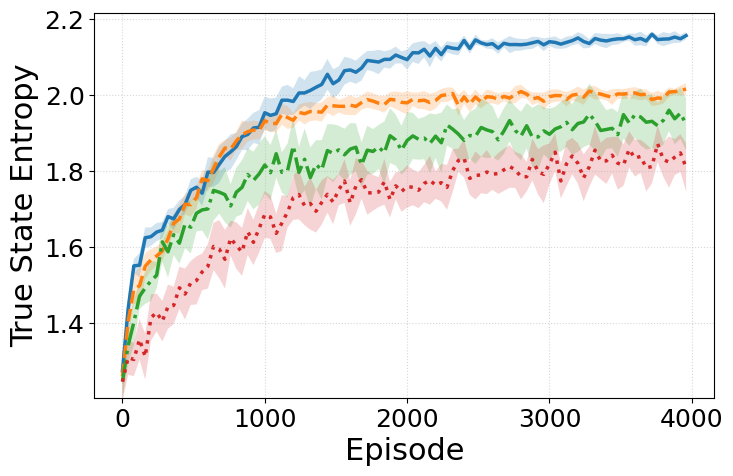}}
        %\hfill
        \subfigure[Env. (a) $(|\mathcal S| = 16)$, deterministic, $0.2$\label{subfig:dynamic4x4}]{\includegraphics[width=0.33\textwidth]{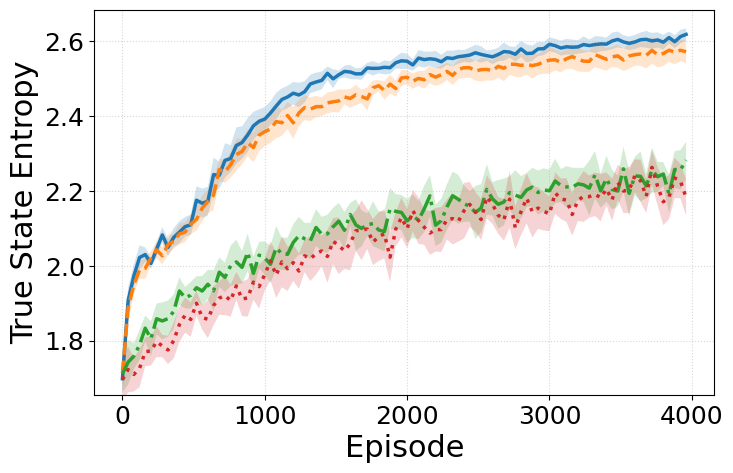}}
    \caption{\emph{True state entropy} obtained by Algorithm 1 with MSE and MBE employing belief averaged policies (BA) and Markovian policies over belief states (B). For each plot, we report a tuple (environment, transition noise, observation variance) where the latter is \emph{not available} (n.a.) when observations are deterministic. For each curve, we report the average and 95\% c.i. over 16 runs. Limited size instances were reported since $|\B|=10^4$ in \ref{subfig:dynamic3x3} and $|\B|=10^5$ in \ref{subfig:dynamic4x4} leading to memory issues in the policies storage. Even in these cases, BA shows higher performances. \label{fig:policy_choice}}
    \end{center}
\end{figure*}

As already described, a plethora of deployable policy classes are possible for addressing MSE in POMDPs. In the main paper, we focused on belief-averaged policies. In Figure~\ref{fig:policy_comparison}, we show how this policy class is superior (or non-worse) to other possible options, being implicitly non-Markovian over observations while being memory efficient. In Figure~\ref{fig:policy_choice}, we show that belief-averaged policies perform better than (direct-parametrization) Markovian policies over belief states, even in the case when the belief states set is manageable in size.

\vspace{1cm}

\newpage

%%%%%%%%%%%%%%%%%%%%%%%%%%%%%%%%%%%%%%%%%%%%%%%%%%%%%%%%%%%%%%%%%%%%%%%%%%%%%%%
%%%%%%%%%%%%%%%%%%%%%%%%%%%%%%%%%%%%%%%%%%%%%%%%%%%%%%%%%%%%%%%%%%%%%%%%%%%%%%%

\end{document}